\documentclass[a4]{article}
\usepackage[letterpaper,top=3cm,bottom=3cm,left=2.5cm,right=2.5cm,marginparwidth=1.75cm]{geometry}
\usepackage{natbib}
\usepackage[hyphens]{url}
\usepackage{lineno,hyperref}
\usepackage{eurosym}
\usepackage{booktabs}
\usepackage{multirow}
\usepackage{graphicx}         
\usepackage{multicol}       
\usepackage{yfonts}
\usepackage{xcolor}
\usepackage{nicefrac}
\usepackage[linesnumbered,ruled,vlined]{algorithm2e}
\usepackage{amsmath,dsfont}
\usepackage{float}
\usepackage{framed}
\usepackage{csquotes}
\usepackage{enumerate}
\usepackage{pdfpages}
\usepackage{mathrsfs}
\usepackage{amssymb,amsmath}
\usepackage{tcolorbox}
\newtcolorbox{mybox}{colback=red!5!white,colframe=red!75!black}
\usepackage[toc,page]{appendix}
\usepackage{mathtools}

\SetCommentSty{mycommfont}
\newtheorem{defi}{Definition}
\SetKwInput{KwInput}{Input}                
\SetKwInput{KwOutput}{Output}              
\newcommand{\LMUTitle}[9]{
  \thispagestyle{empty}
  \vspace*{\stretch{1}}
  {\parindent0cm
   \rule{\linewidth}{.7ex}}
  \begin{flushright}

    \vspace*{\stretch{1}}
    \sffamily\bfseries\Huge
    #1\\
    \vspace*{\stretch{1}}
    \sffamily\bfseries\large
    #2
    \vspace*{\stretch{1}}
  \end{flushright}
  \rule{\linewidth}{.7ex}
  \vspace*{1cm}
  \begin{center}
    \includegraphics[width=2in]{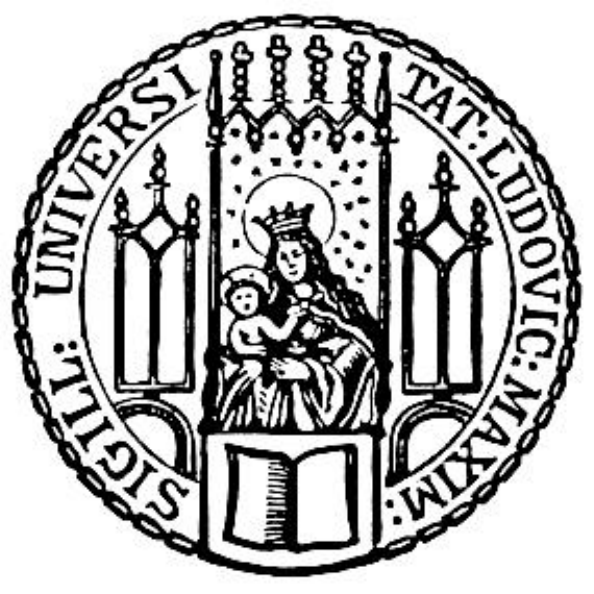}
  \end{center}
  \vspace*{\stretch{1}}
  \begin{center}\sffamily\LARGE{#5}\end{center}

}

\begin{document}

  \LMUTitle
      {\LARGE{Contributions to the\\ Decision Theoretic Foundations of Machine Learning and Robust Statistics under Weakly Structured Information}}               
      {Dr. Christoph Jansen}                       
      {M\"unchen}                             
      {an der Fakult{\"a}t f{\"u}r Mathematik, Informatik und Statistik}                         
      {\normalsize{Kumulative Habilitationsschrift
nach §12(1), Nr. 2 der Habilitationsordnung der\\
Ludwig-Maximilians-Universität München
für die Fakultät für Mathematik, Informatik und Statistik}\\
\normalsize{}}                          
      {27.04.2018}                            
      {Prof.~Dr.~Thomas Augustin}                          
      {Prof.~Dr.}                         
      {Prof.~Dr.} 
      
\allowdisplaybreaks
\newpage 
\thispagestyle{empty}
\quad 
\newpage
\title{Contributions to the\\Decision Theoretic Foundations of Machine Learning and\\ Robust Statistics under Weakly Structured Information}
\date{\small{Kumulative Habilitationsschrift
nach §12(1), Nr. 2 der Habilitationsordnung
der Ludwig-Maximilians-Universität München
für die Fakultät für Mathematik, Informatik und Statistik} \\[1cm] 
vorgelegt von\\[0.3cm]
\large{Dr. Christoph Jansen}}
\thispagestyle{empty}
\maketitle
\thispagestyle{empty}









The present cumulative habilitation thesis is based on ten scientific publications in accordance with §12(1), Nr.~1 and §12(1), Nr.~2 of the Habilitationsordnung mentioned above. In the following sections, the publications submitted in lieu of a Habilitationsschrift are summarized and situated in a broader scientific context in accordance with §12(3) of the Habilitationsordnung. In accordance with §12(2), Sentence 1 of the Habilitationsordnung, I describe my own contributions to the submitted works.\\[9cm]
This habilitation was written under the mentorship of (in alphabetic order):\\[.3cm] Prof.~Dr.~Thomas Augustin (LMU München)\\[.1cm]
Prof.~Dr.~Fabio Cozman (Universidade de S\~{a}o Paulo)\\[.1cm]
Prof.~Dr.~Eyke Hüllermeier (LMU München)

\vspace{2cm}
\newpage 
\thispagestyle{empty}
\quad 
\newpage
\setcounter {page} {1}
\tableofcontents
\newpage
\section*{Contributing Articles} 
\addcontentsline{toc}{section}{Contributing Articles}
The following list contains all publications submitted for the habilitation according to §12(1) Nr.~2 of the Habilitationsordnung. These will be referred to as Contributions 1 to 10 in this thesis. Moreover, Contributions 1, 5 and 10 are the publications that are specifically selected to fulfill §12(1), Nr.~1 of the Habilitationsordnung.


\begin{center}
{\bfseries{A: Decision-Theoretic Foundations}}
\end{center}

\begin{enumerate}
\item Christoph Jansen, Hannah Blocher, Thomas Augustin, and  Georg Schollmeyer (2022): Information efficient learning of complexly structured preferences: Elicitation procedures and their application to decision making under uncertainty. \textit{International Journal of Approximate Reasoning}, 144: 69-91.
\item Jean Baccelli, Georg Schollmeyer, and Christoph Jansen (2022): Risk aversion over finite domains. \textit{Theory and Decision}, 93: 371–397.
\item Christoph Jansen and Thomas Augustin (2022): Decision making with state-dependent preference systems. In: Ciucci, D.; Couso, I.; Medina, J.; Slezak, D.; Petturiti, D.; Bouchon-Meunier, B.; Yager, R.R. (eds): \textit{Information Processing and Management of Uncertainty in Knowledge-Based Systems.} Communications in Computer and Information Science, 1601: 729–742, Springer.
\item Christoph Jansen, Georg Schollmeyer and Thomas Augustin (2023): Multi-target decision making under conditions of severe uncertainty. In:  Torra, V.; Y. Narukawa (eds): \textit{Modeling Decisions for Artificial Intelligence.} Lecture Notes in Computer Science, 13890: 45-57, Springer.
\end{enumerate}

\begin{center}
{\bfseries{B: Machine Learning under Weakly Structured Information}}
\end{center}

\begin{enumerate}
\setcounter{enumi}{4}
 \item Christoph Jansen, Malte Nalenz, Georg Schollmeyer, and Thomas Augustin (2023): Statistical comparisons of classifiers by
generalized stochastic dominance. \textit{Journal of Machine Learning Research}, 24(231): 1–37.
 \item  Hannah Blocher, Georg Schollmeyer, Malte Nalenz and Christoph Jansen (2024): Comparing Machine Learning Algorithms by Union-Free Generic Depth. \textit{International Journal of Approximate Reasoning}, 169: 1-23.
\item Julian Rodemann, Christoph Jansen, Georg Schollmeyer and Thomas Augustin (2023):  In All Likelihoods: Robust Selection of Pseudo-Labeled Data. In: Miranda, E.; Montes, I.; Quaeghebeur, E.; Vantaggi, B. (eds): \textit{Proceedings of the Thirteenth International Symposium on Imprecise Probabilities: Theories and Applications.} Proceedings of Machine Learning Research, 215: 421-425, PMLR.
\item Christoph Jansen, Georg Schollmeyer, Julian Rodemann, Hannah Blocher, Thomas Augustin (2024): Statistical Multicriteria Benchmarking via the GSD-Front. Forthcoming in: \textit{Advances in Neural Information Processing Systems 37 (NeurIPS 2024).} The final version is available at: \url{https://openreview.net/forum?id=jXxvSkb9HD}
\end{enumerate}

\begin{center}
{\bfseries{C: Robust Statistics under Non-Standard Scales of Measurement}}
\end{center}

\begin{enumerate}
\setcounter{enumi}{8}
 \item Hannah Blocher, Georg Schollmeyer, and Christoph Jansen (2022): Statistical models for partial orders based on data depth and formal concept analysis. In: Ciucci, D.; Couso, I.; Medina, J.; Slezak, D.; Petturiti, D.; Bouchon-Meunier, B.; Yager, R.R. (eds): \textit{Information Processing and Management of Uncertainty in Knowledge-Based Systems.} Communications in Computer and Information Science, 1602: 17-30, Springer.
 \item Christoph Jansen, Georg Schollmeyer, Hannah Blocher, Julian Rodemann, and Thomas Augustin (2023): Robust statistical comparison of random variables with locally varying scale of measurement. In: Evans, R.; Shpitser, I. (eds): \textit{Proceedings of the Thirty-Ninth Conference on Uncertainty in Artificial Intelligence.} Proceedings of Machine Learning Research, 216: 941-952, PMLR.
\end{enumerate}







\newpage
\section*{Declaration of the Contributions of the Author}
\addcontentsline{toc}{section}{Declaration of the Author's Contributions}
\iftrue{


In this section, for every publication (called Contribution~1 to Contribution~10 corresponding to the list above) that contributes to the habilitation in the sense of §12(1), Nr.~1  and Nr.~2, I briefly indicate my own contribution to the respective publication.\\

{\noindent {\bfseries{Contribution 1}}} The idea and most parts of the original draft of the paper come from  Christoph Jansen. Georg Schollmeyer worked out most parts of Example~2 (in Section~6) (the simulation part for generating preference systems in this example was done by Hannah Blocher). The executive summary in Section 2  of the paper was worked out by Hannah Blocher and revised by Thomas Augustin. All authors contributed to revising the paper.\\

{\noindent {\bfseries{Contribution 2}}} The paper was mainly drafted by Jean Baccelli. All authors contributed to the main result, Theorem~1. Georg Schollmeyer worked out most parts of the Examples (Example~1 till Example~3), as well as the proofs of the facts to which these examples are related. All authors contributed to revising the paper in several discussion rounds.\\

{\noindent {\bfseries{Contribution 3}}} The idea and the original draft of the paper come from  Christoph Jansen. Thomas Augustin proposed some of the related literature (relating the paper to regret considerations) and wrote parts of Section 1. Both authors contributed to revising the paper in several discussion rounds, especially when incorporating the reviewers' suggestions and comments.\\

{\noindent {\bfseries{Contribution 4}}} The idea and the original draft of the paper come from  Christoph Jansen. All authors contributed to revising the paper in several discussion rounds, especially when incorporating the reviewers' suggestions and comments.\\

{\noindent {\bfseries{Contribution 5}}}
The idea and most parts of the original draft of the paper come from  Christoph Jansen. He also did the programming for the parts related to generalized stochastic dominance and the permutation tests. Georg Schollmeyer supplied Proposition 2 including its proof and its usefulness for the simulation study. He also drafted parts of Section 3.3 and intensively discussed the aspect of incommensurability as well as the role of the hypotheses in the statistical test of Section 4.2. Malte Nalenz designed the simulation study and wrote Sections 5.1 and 6.1. Malte Nalenz did the programming for the parts related to the performance evaluation of the classifiers, for the all-test, the one-test as well as all plots in the paper exept from Figures 5 and 6. Sections 1.2, 5.2 and 6.2 were written jointly by Christoph Jansen and Malte Nalenz. Thomas Augustin wrote Section 1.3. All authors contributed to revising the paper in several discussion rounds.\\

{\noindent {\bfseries{Contribution 6}}} Hannah Blocher developed the idea of ufg depth. She wrote most of the paper. To be more precise: Hannah Blocher wrote all parts except for Section 1.1 (written by Christoph Jansen), Section 1.2 (written by Georg Schollmeyer), Section 2 (written by Christoph Jansen), Section 7.1.2 the comparison to Jansen et al.(2023a) (written by Christoph Jansen and Georg Schollmeyer), the discussion on the order dimension in Section 7.2.1 (written by Georg Schollmeyer) and the conclusion (written by Christoph Jansen and Georg Schollmeyer). Furthermore, Hannah Blocher claimed and proved Lemma~1, Corollary~2, Theorem~3 (1st and 2nd part), Theorem~4 (lower bound), Lemma~6, Corollary~8 and~9 and Theorem~10. Georg Schollmeyer made the claim and proof of the upper bound of Theorem~4. The claim about the connectedness in Theorem~3 (Part 3) was done by Georg Schollmeyer. Hannah Blocher and Georg Schollmeyer provided the proof of Theorem 3. Georg Schollmeyer provided the claim and proof of Theorem~5. The claim of Theorem~7 was done by Georg Schollmeyer and Christoph Jansen. Georg Schollmeyer proved Theorem~7. Hannah Blocher wrote Chapter 7 and implemented the test if a subset is an element of $\mathcal{S}$. Georg Schollmeyer contributed with the implementation of the connectedness property. Malte Nalenz provided the data set, performed the data preparation and wrote Chapter 7.2.1. (except the discussion on order dimension). Georg Schollmeyer, Christoph Jansen and Malte Nalenz supported the analysis with intensive discussions.
Georg Schollmeyer and Christoph Jansen also helped with discussions about the definition of ufg depth and all properties. Malte Nalenz, Christoph Jansen, and Georg Schollmeyer also contributed by providing detailed proofreading and help with the general structure of the paper. Hannah Blocher, Christoph Jansen and Georg Schollmeyer were involved in revising the manuscript and responding to the reviewers.\\

{\noindent {\bfseries{Contribution 7}}} Julian Rodemann developed the main idea of robust PLS extensions that account for model selection, accumulation of error and covariate shift. He drafted and wrote the majority of the paper. Julian Rodemann further implemented robust PLS. He also conceived and conducted the experimental analyses. Christoph Jansen contributed the idea of deploying $\alpha$-cut updating rules for robust PLS. Its regret-based adaption was developed by Julian Rodemann. Christoph Jansen contributed several passages on solving robust PLS problems w.r.t.~generalized stochastic dominance. Georg Schollmeyer and Christoph Jansen also aided with making technical notations more concise. Thomas Augustin, Georg Schollmeyer, Christoph Jansen and Julian Rodemann further contributed by discussions and detailed proof-reading.\\

{\noindent {\bfseries{Contribution 8}}} The idea and most parts of the original draft of the paper come from  Christoph Jansen. Section 5 was jointly written by Julian Rodemann, Christoph Jansen, and Hannah Blocher (in this order). The sufficient condition of a finite VC-dimension and the proof of Theorem 1 are mostly due to Georg Schollmeyer. Corollary 1 was jointly delivered by Georg Schollmeyer and Christoph Jansen. The idea of the dynamic versions of the tests in Section 4 were jointly developed by Christoph Jansen and Georg Schollmeyer. The R code for performing the permutation-based tests in both applications as well as the GitHub repository are due to Hannah Blocher. The idea of using computing time as ordinal performance measure is due to Hannah Blocher. The R code for enabling the benchmark analysis of the PMLB experiments, in particular the hyperparameter tuning and integration of compressed rule ensemble learning (CRE), is due to Julian Rodemann and Georg Schollmeyer. The idea of using feature- and class robustness as ordinal performance measures is due to Georg Schollmeyer. Figures 1, 3, 5, 6, 7 were created by Julian Rodemann, Figures 2 and 4 were created by Christoph Jansen. Thomas Augustin wrote Section 1.1 of the paper.  All authors contributed to revising the paper.\\

{\noindent {\bfseries{Contribution 9}}} Sections 1, 2 and 7 were written jointly by all three authors, where Christoph Jansen contributed most parts of Sections 1 and 7. Hannah Blocher wrote most parts of Sections 3, 4, and 6. She stated and proved Lemma 1 and developed the algorithm for sampling a partial order based on this lemma. The properties  from Definiton~1 were developed and discussed by all authors. The problems of classical methods described in Section~2 were extensively discussed by all authors (and the solution via the concrete scaling method from Section~4 was developed by Hannah Blocher). Georg Schollmeyer wrote most parts of Section~5. He developed the generalizations of the Tukeys depth, the peeling depth and the enclosing depth, as well as the proposal for the weighting within the modified Tukeys depth. All authors contributed to the revision of the paper in several discussion rounds.\\

{\noindent {\bfseries{Contribution 10}}}
The idea and most parts of the original draft of the paper come from  Christoph Jansen. Georg Schollmeyer supplied parts of Proposition 7 and Proposition 8 including parts of the idea for their proof. He also discussed and drafted parts of the aspects related to regularization.  Hannah Blocher wrote Section 8.1 and Appendix B. The R code for performing the permutation-based tests in all three applications as well as the GitHub repository are also due to Hannah Blocher. Julian Rodemann wrote parts of the introduction to Section 6, most parts of Section 8.2 as well as Appendix D. Furthermore, he analyzed and visualized the test results in the paper. The idea for Figure 3 in the main paper as well as Figures 2 and 4 in the appendix was jointly developed by Christoph Jansen and Julian Rodemann. Julian Rodemann proposed the applications on finance and medicine. Thomas Augustin wrote some parts on related literature and parts of the introduction to Section 7. All authors contributed to revising the paper in several discussion rounds.\\

\newpage
}

\fi

\section*{Decision Theory in Machine Learning -- A Brief Introduction}
\addcontentsline{toc}{section}{Decision Theory in Machine Learning -- A Brief Introduction}
\subsection*{On the Shoulders of Giants}
\textit{Decision Theory}, as a discipline originally rather belonging to theoretical economics and philosophy, was discovered early on as a perfectly natural language for addressing problems of mathematical statistics. Among the most prominent examples are \cite{wald}, who reinterprets statistical inference procedures as two-player zero-sum games against nature and solves them by applying classic decision criteria such as minimax, \cite{savage1951theory} who gives an ``informal exposition of it" and adds some ``critical and philosophical remarks", or \cite{Hodges.3}, who propose to evaluate statistical decision functions by mixing their Bayes-risk and their minimax-score. Meanwhile, a decision-theoretic embedding for efficient and elegant solution of statistical problems has become absolutely standard (cf., e.g.,~\cite{witting2013mathematische,berger1985statistical,Berger2000,French2000StatisticalDT,liese} for standard textbooks) and has made it into the standard canon of courses at many statistics faculties. Not without reason Savage's famous theorem, often referred to as the ``crowning glory of decision theory" \citep{k2019}, was first published in a groundbreaking book entitled ``The Foundations of Statistics" \citep{savage1954foundations}.
\\[.2cm]
Presumably, the popularity of decision theory for mathematical statistics is not least due to the fact that its basic model is both simple and expressive and, therefore, allows for a structured treatment of the often math-heavy concepts discussed there, while not losing track in the technical details. Very roughly, the basic model of decision theory is as follows:\footnote{For a more detailed introduction to the basic concepts of decision theory, see Section~\ref{dmuwsi}.} an \textit{agent} (e.g., the statistician) is asked to choose between different \textit{acts} $X$ (e.g., estimators, classifiers or tests) from a known set of acts $\mathcal{G}$. The \textit{consequence} (e.g., the risk or gain) of choosing an act $X$ is not deterministic, but rather depends on which \textit{state of nature} (e.g., parametrization of the model) from a known set $S$ of such states turns out to be the true one. Formally, each act is a mapping $X:S \rightarrow A$, where $A$ is the set of all possible consequences. The \textit{decision problem} $\mathcal{G}$ is then simply a subset (e.g., all unbiased estimators) of the set of all possible acts, i.e., the set $A^S=\{X:S \to A\}$. The agent's goal is then to``solve" the decision problem, i.e., to identify an act, or more gerneral a subset of acts, from $\mathcal{G}$ which is optimal in some predefined sense (e.g., the UMVU estimator or a minimax-optimal test).\\

Moreover, once the embedding of statistics into a decision-theoretic framework has taken place, it is of course possible to draw back on the entire -- sophisticated -- decision-theoretic machinery, with all its possibility and impossibility results, the entire criteriology for optimal decision making as well as a long tradition of axiomatic justifications for this criteriology. In this sense, it can truly be attested that mathematical statistics has benefited immensely from a decision-theoretical perspective, be it in the context of axiomatic justification of Bayesian statistics, a situation-specific criteriology for the evaluation of statistical procedures, the robustification of these procedures, or straightforward extensions to gerneralized uncertainty models such as credal sets or belief functions. Statistical decision theory is a real success story.
\subsection*{What about Machine Learning?}
In view of the popularity of decision-theoretical approaches in classical mathematical statistics just discussed, the question arises as to whether a similar success story can be recorded in machine learning. Here the situation is somewhat more complicated. One the one hand, already in 1988, the seminal paper~\citet{HORVITZ1988247}, co-authored by the current CSO of Microsoft, states: \textit{``Despite their different perspectives, artificial intelligence (AI) and decision~science have common roots and strive for similar goals."} In this same paper, it is argued that a decision-theoretic view is rapidly gaining importance also for machine learning (ML) and artificial intelligence (AI). On the other hand, in 2021 the paper \citet[Section~4]{hüllermeier2021prescriptive} still attests: ``\textit{A decision-theoretic grounding promises to support the systematic development of prescriptive ML methods
in a principled (rather than ad hoc) manner}", thereby emphasizing that decision-theoretic approaches in ML, while expected to be of high benefit for the field, still have not made it into maturity.\\

In my personal view, there is a lot of truth in both perspectives: In contrast to the situation in classical mathematical statistics, in machine learning one searches largely in vain for standard textbooks that take a decision-theoretical perspective. The, nevertheless, many applications of decision-theoretical methods and concepts in machine learning tend to be spread across various research articles, all of which pursue different, rather specific, goals, instead of intending to use decision theory as a common formal framework. A standardized terminology does not seem to exist. \\

Some examples of such applications are given by \citet{UTKIN2007322}, who utilize a decision-theoretic framework to generalized Bayesian learning under imprecise data, \citet{hullermeier2014learning,guillaume2019maximum},  who propose to solve classification problems with set-valued data by drawing back on the minimin solution or the minimax regret solution (both originating from decision-theoretic considerations), \citet{cui}, who utilizes classical decision criteria in problems of machine intelligence, \citet{cd2021}, who propose a robustification of Gaussian discriminant analysis based on Walley's criterion of maximality \citep{Walley.1991}, \citet{shaker}, who propose a measure for uncertainty quantification inspired by the decision-theoretic notion of dominance,\citet{pmlr-v216-rodemann23a}, who express the problem of pseudo-label selection in semi-supervised learning as a decision problem in order to find more robust solution strategies,  or~\citet{schneider}, who review (next to other methods) decision-theoretic approaches to multi-objective hyperparameter optimization. Note that, of course, the small selection of works just listed is exactly this: a selection. Of course, many more approaches exist ranging from Markov decision processes (e.g.,~\citet{DELGADO20111000}) to automated planning under uncertainty (e.g.,~\citet{Trevizan2007PlanningUR}). Additionally, note that there is of course also work in the converse direction, i.e., work on utilizing machine learning approaches for more effective decision making (e.g.,~\citet{mlindt,hanselle}). However, this is not primarily the focus of the present thesis. \\

Finally, it should be noted that also I myself recently co-authored quite a number of research paper in this very field. Next to the articles contributing to the present cumulative habilitation thesis (which are not separately listed here), the following articles can be mentioned here: In~\cite{epub40416}, we investigate first-order stochastic dominance for analyzing partially ordered data, while in \citet{festschrift}, we develop measures for quantifying the stability of Bayes-optimal decisions under severe uncertainty. In \citet{jsa2018}, we introduce the notion of preference systems for sets of consequences with partial cardinal order information, which lay the basis for a number of the contributions in this thesis. In \citet{jsa2017} we give efficient algorithms for computing, e.g., which are optimal with respect to the experience criterion proposed \citet{Hodges.3}, while \citet{JANSEN201849} axiomatically investigates aggregation procedures that account for the homogeneity strucutre of the inputs.
\\[.2cm]
Personally, I am absolutely convinced that a decision-theoretic foundation of modern machine learning is indispensable and that both fields promise to mutually benefit greatly from each other. For this very reason, however, I also think that this promise is still being fulfilled far too little. With the works collected in the appendix of this thesis, I hope I can contribute my (of course very moderate) share on unifying the use of decision-theoretic methods in machine learning. In this context, one specific distinction seems of particular importance for me: A decision-theoretic perspective on machine learning should \textit{not} end with \textit{writing the problem at hand as a decision problem}.\footnote{This would be unfruitful, as almost any scientific problem can be written as a decision making problem under (some kind of) uncertainty.} To the absolute contrary, a decision-theoretic formulation of the problem should only be the first step. What counts is gaining \textit{structural insights} from this translated problem.  While on the one hand, decision theory offers well-motivated criteria for decision making, that can easily be adapted to the considered problem class, on the other hand, the existing axiomatic approaches to decision theory can serve as a guiding principle which definition of optimality should be applied in what kind of situation. While certain data types might be unfamiliar and hard to capture within the traditional machine learning setup, there might be a very natural ways of treating these data once a decision-theoretic embedding has taken place. 
\subsection*{Organization of this Thesis}
This habilitation thesis is \textit{cumulative} and, therefore, is collecting and connecting research that I (together with several co-authors) have conducted over the last few years. Thus, the absolute core of the work is formed by the ten publications listed on page 5 under the name \textit{Contributions 1 to 10}. The complete versions of these articles are attached to this thesis as an appendix, making them as easily accessible as possible for readers wishing to dive deep into the different research projects. The chapters following this section, namely Parts A to C and the concluding remarks, serve to place the articles in a larger scientific context, to (briefly) explain their respective content on a less formal level, and to highlight some interesting perspectives for future research in their respective contexts. Naturally, therefore, the following presentation has neither the level of detail nor the formal rigor that can (hopefully) be found in the papers. The purpose of the following text is to provide the reader an easy and high-level access to this interesting and important research field as a whole, thereby, advertising it to a broader audience. \\

The following text is divided into three parts, A, B and C. \textit{Part A} is entitled \textit{Decision-Theoretic Foundations}. It refers to the first four contributions of this thesis, which can (in a certain sense) be understood as genuinely decision-theoretical. The part starts with an introduction to decision theory under weakly structured information, as it will be consistently understood throughout this thesis. After a high-level content summary of Contributions 1 to 4, part A ends with some exciting directions for (current or) future research. \textit{Part B} is then dedicated to \textit{Machine Learning under Weakly Structured Information}. The part starts with some explanations of how the relevant terms and concepts from Part A can be profitably reinterpreted in the light of machine learning. Following a summary of the content of the articles contributing to this section, Contributions 5 to 8, I will again explain selected promising avenues for future research (and also shed light on some ongoing work). Finally, \textit{Part C} is entitled \textit{Robust Statistics under Non-Standard Scales of Measurement}. After a brief introduction, in which in particular scale-robust stochastic orders and the role of imprecise probabilities for robust statistical testing are discussed, it contains non-technical summaries of the two articles contributing to it, namely Contribution 9 and 10. Also this part ends with a discussion of ongoing and future research. The thesis ends with some concluding remarks.
\begin{figure}[h]
    \centering
    \includegraphics[width=.97\linewidth]{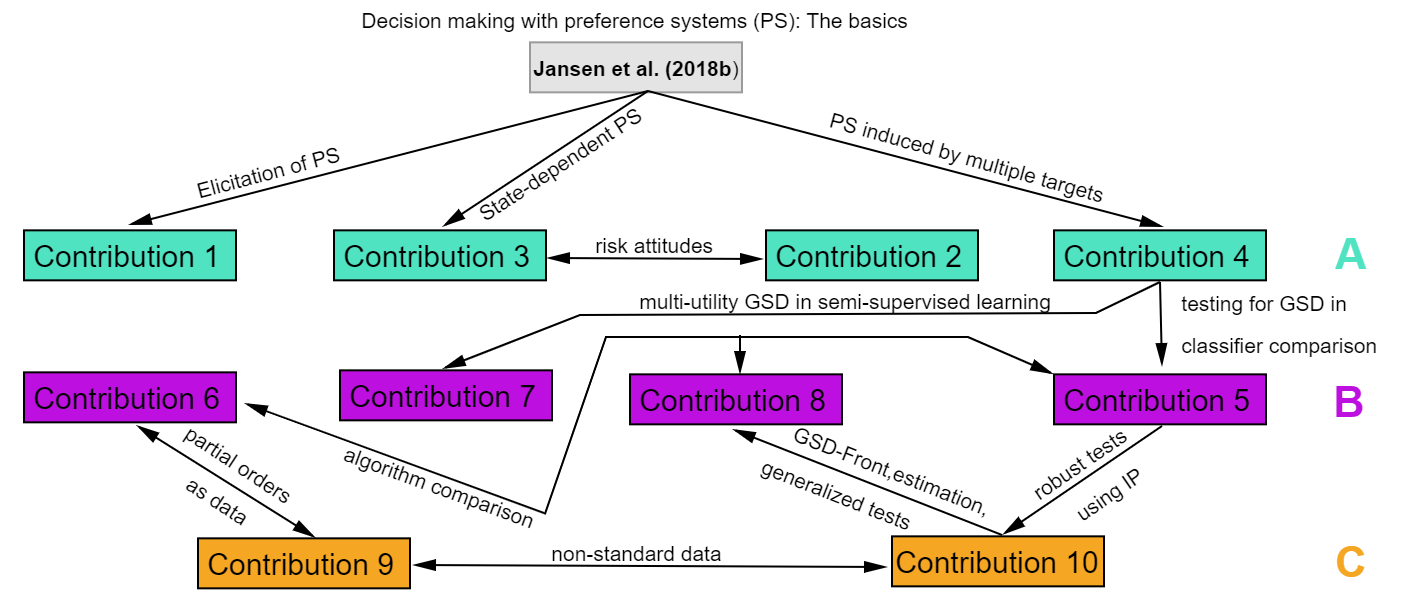}
    \caption{Organization of the contributions to this thesis. The paper highlighted in gray is not part of this habilitation thesis. However, it may be advisable to read it, as it discusses in detail the foundations of many concepts that are important for this thesis, in particular decision making based on preference systems. Apart from this, the graph is to be understood as follows: If a one-sided arrow leads from Contribution A to Contriution B, then B builds (in part) on A. If, on the other hand, A and B are connected by a double arrow, they relate in terms of content, without mathematically (in parts) building directly on each other.}
    \label{example_ps}
\end{figure}
\newpage
\renewcommand\thesection{\Alph{section}}
\section{Decision-Theoretic Foundations} \label{foundations}
\noindent 
\\
\begin{minipage}{0.3\textwidth}

\end{minipage}
\hfill
\begin{minipage}{0.5\textwidth}
\textit{Despite their different perspectives, artificial intelligence (AI) and decision~science have common roots and strive for similar goals.}\\[.2cm] \hfill --------------------------------------(\citet[p.1]{HORVITZ1988247})
\end{minipage}\\[.5cm]

I now come to the first main part of this habilitation thesis: the decision-theoretical foundation. First, a brief introduction to decision theory under weakly structured information, as it is understood here, is given. This is strongly based on the presentation in \cite{jsa2018,jansen2018some}. I will then briefly discuss the four publications that form the basis of this part of the thesis. Finally, I elaborate on promising avenues for future research projects based on the work presented.
\subsection{Decision Making under Weakly Structured Information} \label{dmuwsi}
Decision theory is the study of decision-making under uncertainty, 
 while making the best possible use of all information about preferences, risk attitudes and the underlying uncertainty mechanism. Not least since the groundbreaking works of Ramsey \citep{ramsey1926truth}, De Finetti \citep{definetti}, von Neumann and Morgenstern \citep{neumann}, Savage \citep{savage1954foundations}, and Jeffrey~\citep{jeffrey1990logic}, one paradigm in particular has become established in the last century: \textit{expected utility maximization}. But this paradigm is challenged in situations under weakly structured information settings, as it is based on strong assumptions about both the agent's preferences and the quality of the uncertainty model. I address these situations in the following paragraphs.
\subsubsection{Decision Making under Uncertainty}
The basic formalism of \textit{decision making under uncertainty (DMU)} is as simple as it is expressive: the \textit{agent} (sometimes also called \textit{decision maker}) is asked to choose between different available \textit{acts} $X$ from a known set of acts $\mathcal{G}$. However, the \textit{consequence} of choosing an act $X$ is not (necessarily) deterministic, but rather depends on which \textit{state of the world} from a \textbf{known} set $S$ of such states turns out to be the true one. Formally, each act is a mapping $X:S \rightarrow A$, where $A$ is the set of all possible consequences. The \textit{decision problem} $\mathcal{G}$ is then some fixed subset of the set of all possible acts, i.e., the set $A^S=\{X:S \to A\}$. \\

The agent's goal is to select an optimal (set of) act(s) or, more general, a set of acts for which there is no evidence that one is in some sense superior to the others. This is commonly formalized by a \textit{choice function} $ch:2^{\mathcal{G}} \to 2^{\mathcal{G}}$ satisfying $ch(\mathcal{D}) \subseteq \mathcal{D} $ for all $\mathcal{D} \in 2^{\mathcal{G}}$.\footnote{Note that we explicitly want to allow for empty choice sets. For the original source, see~\citet{sen}.} The sets $ch(\mathcal{D})$ are called \textit{choice sets} and have a slightly different interpretation depending on the quality of the information used to construct the choice function: The \textit{strong view} interprets $ch(\mathcal{D})$ as the set of \textit{optimal} acts from $\mathcal{D}$. The \textit{weak view}, on the other hand, interprets $ch(\mathcal{D})$ as the set of acts from $\mathcal{D}$ that \textit{cannot be rejected} based on the information. More often than not, in this paper we will be concerned with only weakly interpretable choice functions.
\subsubsection{Two Extreme (Yet Classical) Choice Functions}
Usually choice functions are constructed relying on two building blocks: the information about the agent's \textit{preferences}, and the information about the \textit{uncertainty mechanism} generating the states of the world. We will refer to these information bases as $\mathcal{I}_1$ and $\mathcal{I}_2$, respectively.\footnote{Note that in most parts of this thesis we assume the information bases $\mathcal{I}_1$ and $\mathcal{I}_2$ to be describable \textit{independently} of each other. An exception is given by Contribution 3, which assumes the agent's preferences to be dependent on the states of nature of the decision problem under consideration. Classic theoretical sources on state-dependent preferences and utility functions are \cite{karni2013decision}~or~\cite{ssk1,ssk2}, but many more exist. \cite{jean,baccelli2022expected} provide a modern reappraisals.} For further reference, we will briefly discuss two popular choice functions at this point:\footnote{In order to preserve the non-technical character of this introduction, we always assume (now and in the following) that all occurring expectations exist and are finite. The exact technical assumptions can be found in the relevant publications contributing to this thesis.\label{nontechnical}}\\

\textbf{Expected Utility (EU):} If the infomation in $\mathcal{I}_1$ allows to describe the uncertainty about the states by a (subjective) probability $\pi$ on $S$, and the information in $\mathcal{I}_2$ allows for expressing the agent's preferences by a cardinal \textit{utility function} $u:A \to [0,1]$, then we can set
\begin{equation}\label{exut}
ch_{u,\pi}(\mathcal{D})=\Bigl\{Y \in \mathcal{D}: \mathbb{E}_\pi(u \circ Y) \geq \mathbb{E}_\pi(u \circ X) \text{ for all }X \in \mathcal{D}\Bigr\},
\end{equation}
for all $\mathcal{D} \subseteq \mathcal{G}$ and choose that acts from $\mathcal{D}$ that \textit{maximize expected utility}. The choice function $ch_{u,\pi}$ then obviously allows for a strong interpretation: all acts $ch_{u,\pi}(\mathcal{D})$ produce coinciding expected utilities, which are strictly greater than all expected utilities of acts in $\mathcal{D} \setminus ch_{u,\pi}(\mathcal{D})$. Consequently, the agent is \textit{indifferent} with among the equally optimal acts contained in $ch_{u,\pi}(\mathcal{D})$.\\

\textbf{First-Order Stochastic Dominance (FSD):} If the infomation in $\mathcal{I}_1$ allows to describe the uncertainty about the states by a (subjective) probability $\pi$ on $S$, and the information in $\mathcal{I}_2$ consists only in a preorder $\succsim$ on $A$ (see Section~\ref{basics}), we can set
\begin{equation} \label{FSD}
ch_{\succsim,\pi}(\mathcal{D})=\Biggl\{Y \in \mathcal{D}: \nexists X\in \mathcal{D} ~\text{ s.t. }~ \begin{array}{lr} \mathbb{E}_\pi(u \circ X) \geq \mathbb{E}_\pi(u \circ Y) ~~~\text{for all} ~~~~u \in \mathcal{U}_{\succsim}\\ \mathbb{E}_\pi(u \circ X) > \mathbb{E}_\pi(u \circ Y) ~~~\text{for some} ~u \in \mathcal{U}_{\succsim}
\end{array}\Biggr\}  
\end{equation}
for all $\mathcal{D} \subseteq \mathcal{G}$, where $\mathcal{U}_{\succsim}$ is the set of all $\succsim$-isotone $u:A \to [0,1]$. Thus, we choose all acts that are \textit{not excluded by every compatible EU-maximizer}. The choice function $ch_{\succsim,\pi}$ then obviously allows only for a weak interpretation: among all acts in $ch_{\succsim,\pi}(\mathcal{D})$ the agent is either \textit{indifferent} or considers them \textit{incomparable}. The acts in $\mathcal{D} \setminus ch_{\succsim,\pi}(\mathcal{D})$ are all strictly stochastically dominated by at least one act from $ch_{\succsim,\pi}(\mathcal{D})$.
\subsubsection{Some Basic Definitions} \label{basics}
Before we can turn to decision situations under less extreme information scenarios, their formalization requires a few basic mathematical concepts. We will now summarize these, guided by~\citet{chambers2016revealed}.\\

A binary relation $R$ on a set $M \neq \emptyset$ is a subset of the Cartesian product of $M$ with itself, i.e.~$R \subseteq M \times M$. $R $ is called \textit{reflexive}, if $(a,a) \in R$, \textit{transitive}, if $(a,b),(b,c) \in R\Rightarrow(a,c) \in R$, \textit{antisymmetric}, if $(a,b),(b,a) \in R\Rightarrow a=b$, \textit{complete}, if $(a,b) \in R$ or $(b,a) \in R$ (or both) for arbitrary elements $a,b,c \in M$. A \textit{preference relation} is a binary relation that is complete and transitive; a \textit{preorder} is a binary relation that is reflexive and transitive; a \textit{linear order} is a preference relation that is antisymmetric; a \textit{partial order} is a preorder that is antisymmetric. If $R$ is a preorder, we denote by $P_R \subseteq M \times M$ its \textit{strict part} and by $I_R \subseteq M \times M$ its \textit{indifference part}, defined by $ (a , b) \in P_{R} \Leftrightarrow (a , b) \in R \wedge (b , a) \notin R,$
and $(a , b) \in I_{R} \Leftrightarrow (a , b) \in R \wedge (b , a) \in R$.\\

This leads us to the central ordering structure under consideration in most parts of the present thesis, namely \textit{preference systems}. These formalize the idea of agents with partial ordinal and partial cardinal preferences and are discussed, e.g., in~\cite{jsa2018}.
\begin{defi}\label{ps}
Let $A \neq \emptyset$ be a set, $R_1 \subseteq A \times A$ a preorder on $A$, and
 $R_2 \subseteq R_1 \times R_1$ a preorder on $R_1$. The triplet $\mathcal{A}=[A, R_1 , R_2]$ is then called a \textbf{preference system} on $A$. We call $\mathcal{A}$ \textbf{bounded}, if there exist $a_*,a^* \in A$ such that $(a^*,a) \in R_1$, and $(a,a_*) \in R_1$ for all $a \in A$, and $(a^*,a_*) \in P_{R_1}$. Moreover, the preference system $\mathcal{A}'=[A', R_1' , R_2']$ is called \textbf{subsystem} of $\mathcal{A}$ if $A' \subseteq A$, $R_1'\subseteq R_1$, and $R_2'\subseteq R_2$. In this case, we call $\mathcal{A}$ a \textbf{supersystem} of $\mathcal{A}'$.
\end{defi}
While the relation $R_1$ formalizes the available ordinal information, i.e. information about the arrangement of the elements of $A$, the relation $R_2$ describes the cardinal part of the information in the sense that pairs standing in relation are ordered with respect to the intensity of the relation. Thus, intuitively speaking, the set $A$ is locally almost cardinally ordered on subsets where $R_1$ and $R_2$ are very dense, while on subsets where $R_2$ is sparse or even empty, locally at most an ordinal scale of measurement can be assumed. A graphical example that illustrates the intuition underlying preference systems can be found in Figure~\ref{example_ps}.
\begin{figure}
    \centering
    \includegraphics[width=.65\linewidth]{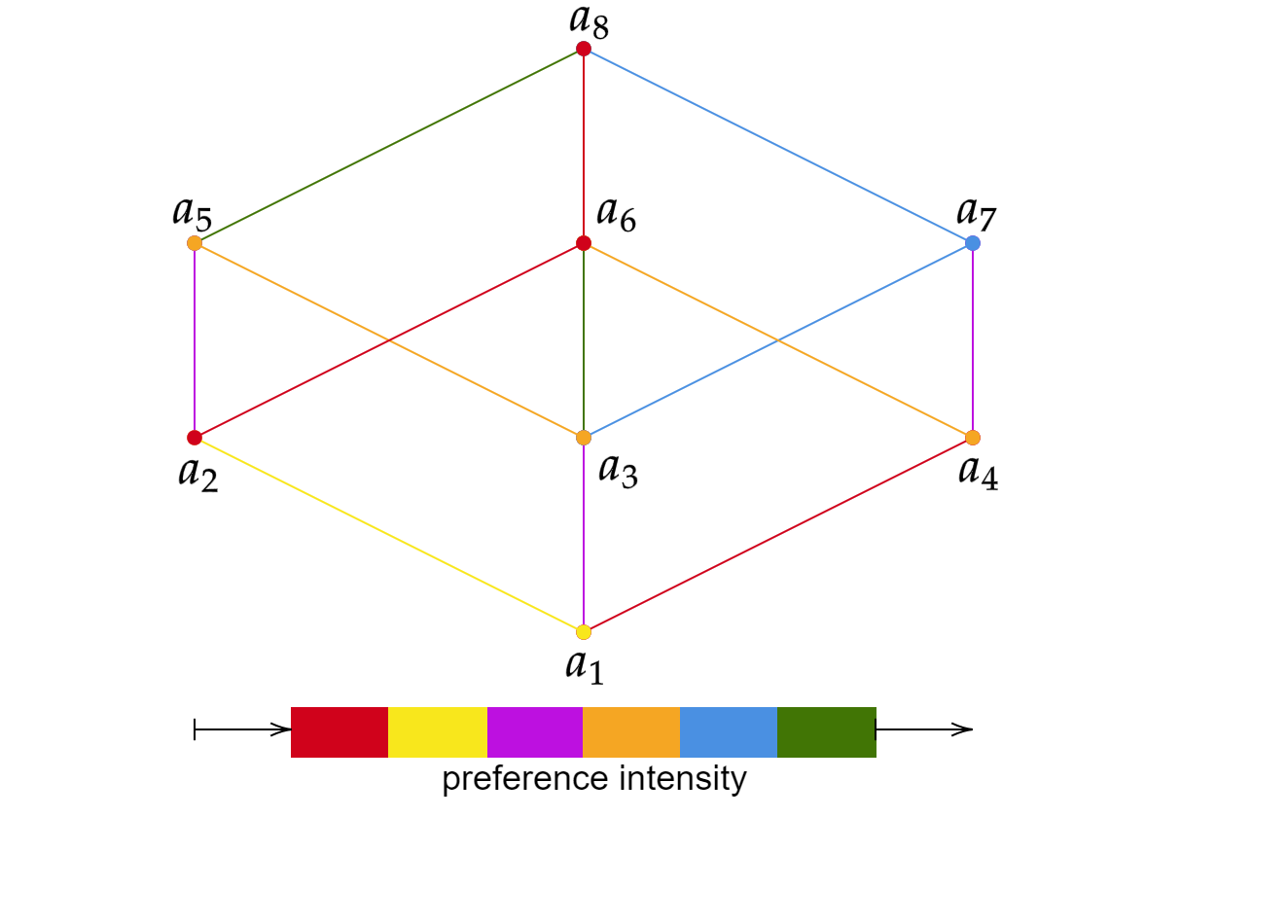}
    \caption{The image shows the Hasse graph of the relation $R_1$ of an exemplary preference system $\mathcal{A}$ on the finite consequence set $A = \{a_1, \dots , a_8\}$. The edges of the Hasse graph are sorted by intensity according to the color spectrum below. This symbolizes the relation $R_2$ of the exemplary preference system. For example, every representation $u \in \mathcal{U}_{\mathcal{A}}$ must satisfy that $u(a_7)-u(a_3)> u(a_4)-u(a_1)$ applies, as the blue preference is more intense than the red preference.}
    \label{example_ps}
\end{figure}
To ensure that $R_1$ and $R_2$ are compatible, we use a consistency criterion for preference systems relying on the idea that both relations $R_1$ and $R_2$ should be simultaneously representable by the same utility function.
\begin{defi} \label{consistency}
The preference system $\mathcal{A}=[A, R_1 , R_2]$ is \textbf{consistent} if there exists a \textbf{representation} $u:A \to \mathbb{R}$ such that for all $a,b,c,d \in A$ we have:
\begin{itemize}\itemsep1.5mm
\item[i)] If we have that $ (a , b) \in R_1$, then it holds that $u(a) \geq u(b)$, where equality holds if and only if $(a,b)\in I_{R_1}$.
\item[ii)] If we have that $((a , b),(c,d)) \in R_2$, then it holds that $u(a) -u(b) \geq u(c)-u(d)$, where equality holds if and only if $((a,b),(c,d))\in I_{R_2}$.
\end{itemize}
The set of all representations of $\mathcal{A}$ is denoted by $\mathcal{U}_{\mathcal{A}}$. 
\label{consistent}
\end{defi}
In certain situations, for example in the context of regularization of preference systems as discussed in publication 10 of this paper, we are only interested in the set of all normalized representations of a preference system. This consideration is the subject of the following definition.
\begin{defi} \label{granularity}
Let $\mathcal{A}=[A, R_1 , R_2]$ be a consistent and bounded preference system with $a_*,a^*$ as before. Then 
$$\mathcal{N}_{\mathcal{A}}:= \Bigl\{u  \in \mathcal{U_{\mathcal{A}}}: u(a_*)=0 ~\wedge ~ u(a^*)=1 \Bigl\}$$
is called the \textbf{normalized representation set} of $\mathcal{A}$. Further, for $\delta \in [0,1)$, we denote by $\mathcal{N}^{\delta}_{\mathcal{A}}$ the set of all $u \in \mathcal{N}_{\mathcal{A}}$ with
 $$ u(a)-u(b) \geq \delta ~~~\wedge ~~~ u(c)-u(d) -u(e) + u(f) \geq \delta $$ 
 for all $(a,b) \in P_{R_1}$ and for all $((c,d),(e,f)) \in P_{R_2} $. We call $\mathcal{A}$ \textbf{$\delta$-consistent} if $\mathcal{N}^{\delta}_{\mathcal{A}} \neq \emptyset$.
\end{defi}
\subsubsection{A Choice Functions Between the Extremes} \label{extremes}
Building on the conceptual and mathematical premises just repeated, I now discuss the framework of \textit{decision theory under weakly structured information} that is a central theme throughout this work and the publications contributing to it. The idea here is to enable a more general and thus more flexible modeling for both information sources $\mathcal{I}_1$ (i.e.~the preference information) and $\mathcal{I}_2$ (i.e.~the uncertainty information).\\

We have just laid the foundation for a more flexible modeling of $\mathcal{I}_1$: Instead of directly assuming a cardinal utility function on the consequences in $A$, or modeling purely ordinally via preference relations, the agent may now express their preferences via a \textit{preference system.} In this way, as already indicated above, the agent can also express partially cardinal preference structures in the sense that information about the \textit{intensity of preference} between certain consequences is known, whereas there may be incomparability or only a ranking between other consequences. In this case, the entire set $\mathcal{U}_{\mathcal{A}}$ (or sometimes $\mathcal{N}_{\mathcal{A}}$) of representations of the preference system serves as a model for the agent's preferences, instead of just a single cardinal utility function.\\

Next, let's turn to the relaxation of $\mathcal{I}_2$, i.e., the information about the process that generates the states of nature. While $\mathcal{I}_2$ is classically (im- or explicitely) assumed to be describable by a single probability measure, \textit{imperfect probabilistic information} will be present in many in real-world applications, e.g., in the form of constraints on the probabilities of certain events or, more generally, on the expectations of certain random variables. To describe this kind of generalized uncertainty, the theory of \textit{imprecise probabilities} as developed in \citet{williams2007notes} (specifically as \textit{conditional previsions}), \cite{l1974} (specifically \textit{credal sets}), \cite{kofler} (specifically \textit{linear partial information}), \cite{WILLIAMSON199089} (specifically as \textit{p-boxes}), \cite{Walley.1991} (specifically \textit{lower previsions}), \citet{w2001} (specifically \textit{interval probability}) is perfectly suitable.\footnote{For  recent introductions to the theory  see~\citet{augustin2014introduction} and \citet{Bradley:2019}.} It should be noted here that the term imprecise probabilities is actually an umbrella term for many different generalized uncertainty theories. I restrict myself to  convex finitely generated credal sets.
\begin{defi}\label{credset}
A \textit{finitely-generated credal set} on a measurable space $(S ,\sigma(S))$ is a set 
\begin{equation*}
\mathcal{M}=\Bigr\{ \pi \in \mathcal{P}: \underline{b}_\ell \leq \mathbb{E}_\pi(f_\ell) \leq \overline{b}_\ell \text{ for } \ell=1, \dots, r\Bigl\}
\end{equation*}
with $\mathcal{P}$ the set of all probabilities on $(S ,\sigma(S))$, $f_1, \dots , f_r: S \to \mathbb{R}$  bounded and measurable, and $\underline{b}_\ell \leq \overline{b}_\ell$ their lower and upper expectation bounds.
\end{defi}
%
Credal sets obviously represent a genuine generalization of classical probability theory: If the agent has perfect probabilistic knowledge about the states in the form of a probability measure $\pi$, then the credal set can simply be the set containing only this measure, i.e. $\mathcal{
M}=\{\pi\}$ can be chosen. If, on the other hand, there is no probabilistic knowledge at all, this can also be modeled using credal sets: Here you simply choose the set of all possible probability measures, i.e. $\mathcal{M}= \mathcal{P}$. What makes credal sets particularly attractive, however, is that they can be used to model all information states between these two extreme scenarios: One simply collects all classical probability measures in a set that do not contradict the existing (partial) probabilistic knowledge. An illustration of an exemplary credal set can be found in Figure~\ref{example_pol}.\\

Using the just discussed relaxations of the models for the information sources underlying the decision problem (preference systems for $\mathcal{I}_1$ and credal sets for $\mathcal{I}_2$), we can now define customized choice functions for this situation as well. Much work has been done in the literature on the case where the information source $\mathcal{I}_1$ was replaced by an imprecise probabilistic model,\footnote{See, e.g.,~\cite{troffaes_ijar} for a survey or \cite{l1974,Walley.1991,festschrift} for original sources. Note that there is also quite an amount of literature on computation for that case, see, e.g.,~\cite{autkin,kikuti,compitip,jsa2017,maua}.} while the information source $\mathcal{I}_2$ was typically left untouched.\footnote{An exception is \cite{pivato}, who axiomatically discusses multi-utility representations for incomplete difference preorders.} A recent work on choice functions under generalization of both information sources simultaneously is given by~\cite{jsa2018}. We focus here on only one choice function, namely the one induced by generalized stochastic dominance (GSD), which is closely related to the relation $R_{\forall\forall}$ discussed in~\cite[p. 123]{jsa2018}. The idea behind GSD is very simple: An act is at least as good as a competitor, if it has at least as high expected utility with respect to any combination of utility representing the preference system and probability belonging to the credal set. More formally, we arrive at:
\begin{figure}
    \centering
    \includegraphics[width=.42\linewidth]{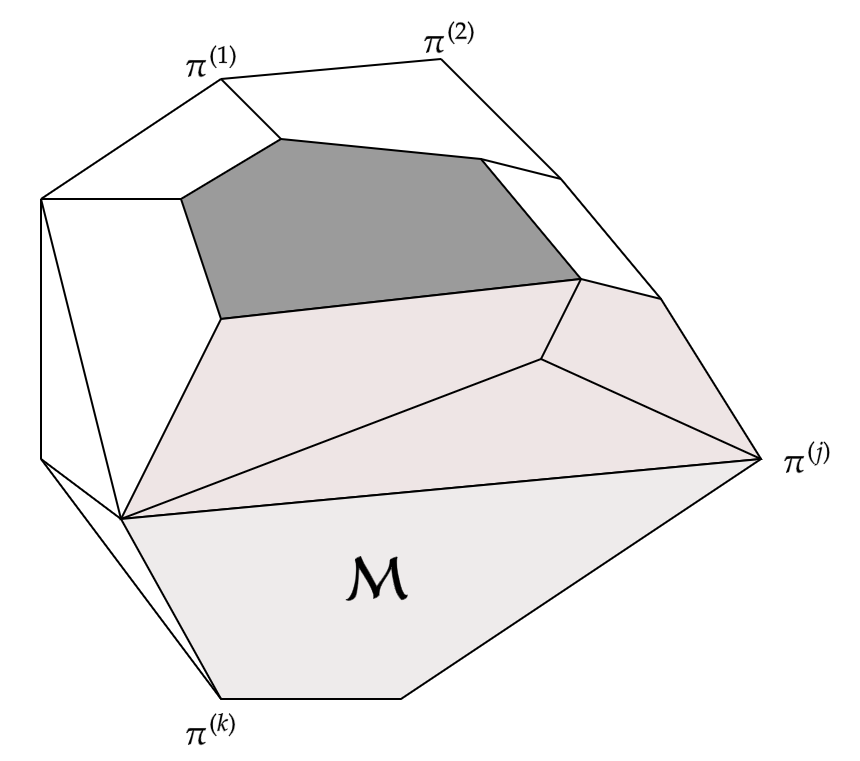}
    \caption{The figure shows an exemplary credal set on a finite set of states $S=\{s_1,s_2,s_3\}$. You can see directly that this can be identified (for finite $S$) with a convex polyhedron. We repeatedly take advantage of this fact in some of the publications to solve various problems using (mixed-integer) linear programming.}
    \label{example_pol}
\end{figure}
\begin{defi}\label{GSD}
Let $\mathcal{A}$ be a consistent preference system and let $\mathcal{M}$ be a finitely generated credal set. We then say an act $Y$ is \textbf{$(\mathcal{A},\mathcal{M})$-dominated} by an act $X$ if 
$$\mathbb{E}_{\pi}(u \circ X) \geq \mathbb{E}_{\pi}(u \circ Y)$$
 for all $u \in \mathcal{U}_{\mathcal{A}}$ and for all $\pi \in \mathcal{M}$.\footnote{Note that, again, we leave measurability issues aside, compare Footnote~\ref{nontechnical}. The detailed and technical rigorous definitions can be found in the respective contributions.} The induced binary relation among acts is denoted by $R_{(\mathcal{A},\mathcal{M})}$ and called \textbf{generalized stochastic dominance (GSD)}.
\end{defi}
One easily sees that $R_{(\mathcal{A},\mathcal{M})}$ defines a preorder on the set of acts that, in general, will possess a non-trivial imcomparability part. Thus, it naturally induces a choice function $ch_{R_{(\mathcal{A},\pi)}}$ on a fixed decision problem $\mathcal{G}$, by taking for every inserted $\mathcal{D} \subseteq \mathcal{G}$ the undominated acts from $\mathcal{D}$. Formally, this reads as
\begin{equation} \label{GSD1}
ch_{R_{(\mathcal{A},\mathcal{M})}} (\mathcal{D}):=\Bigl\{X \in \mathcal{D}: \nexists Y\in \mathcal{D} \text{ such that }(Y,X) \in P_{R_{(\mathcal{A},\mathcal{M})}}  \Bigr\}.    
\end{equation}
$\mathcal{D} \subseteq \mathcal{G}$, or, if we rather intend to emphasize the structural similarity to the choice function from Equation~(\ref{FSD}), then equivalently also as
\begin{equation}\label{GSD2}
ch_{R_{(\mathcal{A},\mathcal{M})}} (\mathcal{D})=\Biggl\{Y \in \mathcal{D}: \nexists X\in \mathcal{D} ~\text{ s.t. }~ \begin{array}{lr} \mathbb{E}_\pi(u \circ X) \geq \mathbb{E}_\pi(u \circ Y) ~~~\text{for all} ~~~~u \in \mathcal{U}_{\mathcal{A}}\\ \mathbb{E}_\pi(u \circ X) > \mathbb{E}_\pi(u \circ Y) ~~~\text{for some} ~u \in \mathcal{U}_{\mathcal{A}}
\end{array}\Biggr\}.  
\end{equation}
A comparison of the choice functions defined in (\ref{FSD}) and (\ref{GSD2}) now directly makes some connections clear: The choice function (\ref{GSD2}) is a generalization of the classical first-order stochastic dominance (\ref{FSD}), which additionally allows to include partial cardinal information and partial probabilistic knowledge in the analysis. Consequently, (\ref{GSD2}) also coincides with (\ref{FSD}) if no additional cardinal information is available, i.e.~$R_2$ of the underlying preference system is the trivial preorder and if the credal set is $\mathcal{M}=\{\pi\}$, i.e., consists of only one probability measure. Note, moreover, that under strong assumptions on the underlying preference system, the choice function even reduces to classical expected utility as given in Equation~(\ref{exut})~\citep[Chapter 4]{krantz}.\footnote{Note that generalizing stochastic dominance to settings under generalized uncertainty models is a much-studied problem, including prominent works like \citet{denoeux,montemir,miranda,COUSO2024109113}. The conceptual difference is that the approach studied here also allows for including partial cardinal information on preferences.}
%
\subsection{Contributions for Part A}
I will now briefly summarize the four publications forming the core of Part A of the present work. The order in which the publications are discussed is chronological by date of publication, starting with:
\begin{mybox}
\begin{center}
    {\bfseries{Contribution 1}}
\end{center}
Christoph Jansen, Hannah Blocher, Thomas Augustin, and  Georg Schollmeyer (2022): Information efficient learning of complexly structured preferences: Elicitation procedures and their application to decision making under uncertainty. \textit{International Journal of Approximate Reasoning}, 144: 69-91.\\

\begin{center}
    {\bfseries{Original Abstract}}\\
\end{center}

\textit{In this paper we propose efficient methods for elicitation of complexly structured preferences and utilize these in problems of decision making under (severe) uncertainty. Based on the general framework introduced in \cite{jsa2018}, we now design elicitation procedures and algorithms that enable decision makers to reveal their underlying preference system (i.e. two relations, one encoding the ordinal, the other the cardinal part of the preferences) while having to answer as few as possible simple ranking questions. Here, two different approaches are followed. The first approach directly utilizes the collected ranking data for obtaining the ordinal part of the preferences, while their cardinal part is constructed implicitly by measuring the decision maker's consideration times. In contrast, the second approach explicitly elicits also the cardinal part of the decision maker's preference system, however, only an approximate version of it. This approximation is obtained by additionally collecting labels of preference strength during the elicitation procedure. For both approaches, we give conditions under which they produce the decision maker's true preference system and investigate how their efficiency can be improved. For the latter purpose, besides data-free approaches, we also discuss ways for statistically guiding the elicitation procedure if data from elicitations of previous decision makers is available. Finally, we demonstrate how the proposed elicitation methods can be utilized in problems of decision under (severe) uncertainty. Precisely, we show that under certain conditions optimal decisions can be found without fully specifying the preference system.}
\end{mybox}
\bigskip
The first article deals with the \textit{elicitation}, i.e. the systematic querying, of preference systems (in the sense of Definition~\ref{ps}). The study of elicitation schemes of preference and utility structures has a long tradition, both in economics (e.g.,~\cite{suppes,abdel,fischhoff}) and in psychologie/philosophy (e.g.,~\cite{walsh,eugene,baccelli2016choice}). At least since the emergence of streaming platforms such as Netflix or Amazon Prime and the preference-based recommendation algorithms associated with them, the topic has also become increasingly and rapidly important in machine learning (e.g.,~\cite{haddaway,abbas,gerding}).  Our focus here is on \textit{information efficiency}, in the sense that the person whose preferences are to be elicited should have to answer as few and as simple questions as possible, which ideally only depend on the specific factual context. The special feature here is that the aim is \textit{not} to elicit the entire preference system as quickly as possible, but rather \textit{only those parts} of this system that are necessary to evaluate a choice function in a decision problem under weakly structured information. In contrast to many common approaches, we thus do not directly query the utility function, but successively shrink the set of utility functions still in question by querying the underlying preference system. \\

But first things first: Denote by $\mathcal{A}^*=[A,R_1^*,R_2^*]$ the agent's true (but unknown) preference system, where throughout the relations $R_1^*$ and $R_2^*$ are assumed to be reflexive, and the set $A=\{a_1 , \dots , a_n\}$ is assumed to be finite. We want to elicit $\mathcal{A}^*$ in an information efficient manner. Information efficiency can here be formalized as follows: The agent should be asked fewest possible simple ranking questions concerning $R_1^*$, whereas $R_2^*$ should be obtained implicitly from the elicitation process. For doing so, the paper proposes two different procedures:\\

\textbf{Time Elicitation:} The intuition behind time elicitation is that the strength of the preference between two consequences \textit{decreases} in the time that ranking the two consequences takes. In particular, whenever the agent strictly prefers $a_i$ to $a_j$, we measure the \textit{consideration time} $t_{ij}$ the agent needs for ranking the two consequences. If the agent judges the pair to be incomparable, we set the consideration time to $c_{\infty}>  \max\{t_{pq}: (a_p,a_q) \in P_{R_1^*}\}$, whereas if $a_i$ and $a_j$ are incomparable for the agent the consideration time is set to $0$. We then successively construct an approximation $R_1$ of $R_1^*$ by adding $(a_i , a_j)$ if $t_{ij}>0$. Afterwards, we add the pair of pairs $((a_i,a_j),(a_k,a_l))$ to $R_2$ if $t_{kl}-t_{ij}\geq 0 ~\wedge~t_{ij}>0$.\\

\textbf{Label Elicitation:} In label elicitation the agent assigns each pair of consequences $(a_i , a_j) \in A \times A$ one label from the set $\mathcal{L}_r:= \{\mathbf{n}, \mathbf{c},0,1 , \dots , r\}$, where $r \in \mathbb{N}$ is an integer. The higher the label $\ell_r^{ij}$ of $(a_i , a_j) \in A \times A$ is, the stronger is the agent's strict preference of $a_i$ over $a_j$. If the label $\mathbf{n}$ is assigned, this means that $a_i$ and $a_j$ are incomparable, whereas the label $0$ is interpreted as indifference.  If the label $\mathbf{c}$ is assigned, this means that $a_i$ is strictly preferred to $a_j$, but no statement about the intensity of preference is possible. The collected labels are utilized to successively build up a preference system: Whenever $\ell_r^{ij} \in \mathcal{L}_r\setminus \{\mathbf{n},0\}$, we add the pair $(a_i,a_j)$ to our relation $R_1$. If $\ell_r^{ij}=0$, we add both pairs $(a_i,a_j)$ and $(a_j,a_i)$ to our relation $R_1$, whereas if $\ell_r^{ij}=\mathbf{n}$ the relation $R_1$ remains unchanged. Afterwards, we successively pick pairs of pairs $(a_i , a_j)$, $(a_k , a_l) \in R_1$ and add $((a_i , a_j),(a_k , a_l))$ to our relation $R_2$ if and only if $\ell_r^{ij}>\ell_r^{kl}$ or $\ell_r^{ij}=\ell_r^{kl} =0$.\\

Both procedure produce a preference system $\mathcal{A}=[A,R_1,R_2]$. Importantly, note that without further assumptions neither of those preference systems does have to coincide or even be a sub-system of the agent's true preference system $\mathcal{A}^*=[A,R_1^*,R_2^*]$. For this reason, in the paper, we each give a set of sufficient conditions under which the procedures converge to the true preference system, see \citet[Proposition 1]{JANSEN202269} for time elicitation and \citet[Proposition 4]{JANSEN202269} for label elicitation. Moreover, we propose more efficient versions of the basic versions of the above procedures: For time elicitation, we show that the procedure can be drastically improved if the agent to be elicited has transitive preferences \citep[Algorithhm 1]{JANSEN202269}. In contrast, for label elicitation, we propose a hierarchical modification that is guaranteed to converge to the true preference systems already for very moderate numbers of labels $r \geq 2$  \citep[Proposition 5 and Algorithm 2]{JANSEN202269}.\\

The rest of the paper is then devoted to the decision-making situation \textit{under uncertainty}. More precisely, the situation considered so far is modified as follows: We assume the agent's preferences on $A$ are adequately described by the preference system $\mathcal{A}^*=[A,R^*_1,R^*_2]$, where still $A=\{a_1, \dots ,a_n\}$ is a finite set of consequences. However, the consequence that a specific decision produces now depends on which state of nature from $S=\{s_1 , \dots ,s_m\}$ occurs. The agent then is faced with a finite set of acts $\mathcal{G} =\{X_1 , \dots , X_k\} \subseteq A^S$ out of which it may be chosen. Moreover, we assume there is also information on the mechanism generating the states $s \in S$. This information is assumed to be characterized by a polyhedral \textit{credal set} $\mathcal{M}$ of probability measures on the set $S$ in the sense of Definition~\ref{credset}. \\

\begin{figure}[h]
    \centering
    \includegraphics[width=.4\linewidth]{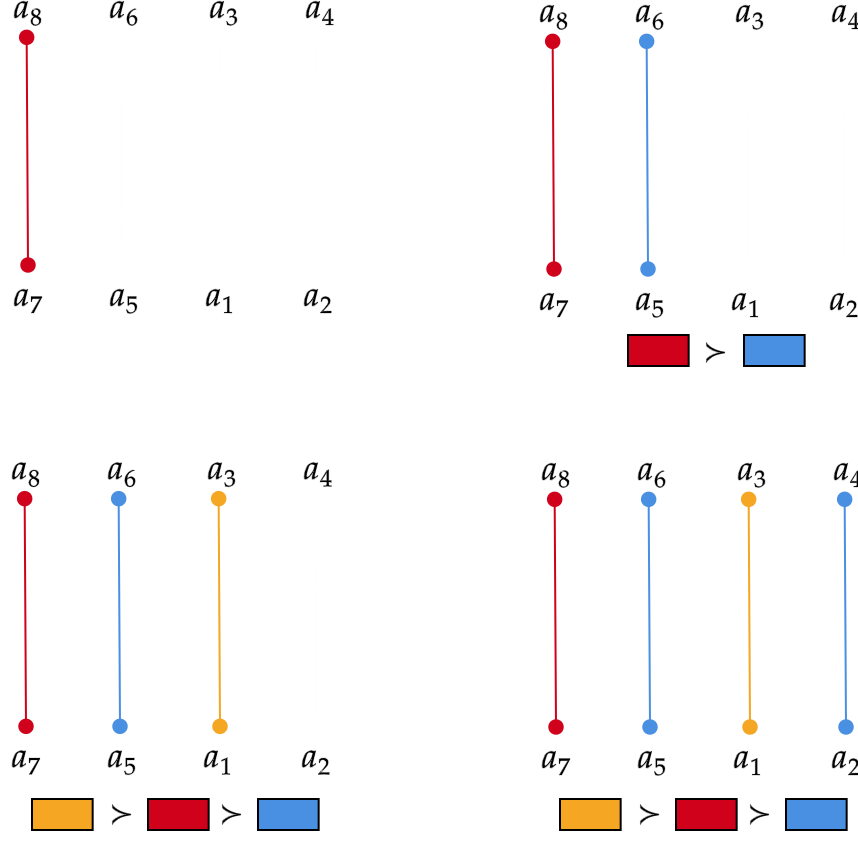}
    \caption{Label eliciation with three labels (yellow $=3$ $>$ red $=2$ $>$ blue $=1$). Consider a decision problem with $A=\{a_1, \dots , a_8\}$, $S=\{s_1 , \dots , s_4\}$, $\mathcal{G}=\{X_1, X_2\}$, $X_1(S)=(a_8,a_5,a_2,a_3)$,  $X_2(S)=(a_7,a_6,a_4,a_1)$, and $\mathcal{M}=\{\pi: \pi(\{s_1\}) \geq \pi(\{s_2\}) \geq\pi(\{s_4\}) \geq \pi(\{s_3\})\}$. After asking four ranking questions (see above the single elicitation steps), we can conclude 
$4\cdot(\mathbb{E}_{\pi}(u \circ X_1)-\mathbb{E}_{\pi}(u \circ X_2)) = (u_8-u_7)-(u_6-u_5)+(u_3-u_1)-(u_4-u_2) >0,$ for every $u \in \mathcal{U}_{\mathcal{A}^*}$. Hence, we can conclude taht $X_1$ is optimal after asking four simple ranking questions. (The example is similar to Example 1 in Contribution 1.)}
    \label{example_eli}
\end{figure}

The task of elicitation is now slightly modified: Instead of just wanting to shrink the set $\mathcal{U}_{\mathcal{A}^*}$ of compatible utility functions as efficiently as possible -- as in the situation under certainty before -- we now want to elicit in such a way that a previously known choice function (based on $\mathcal{A}^*$ and $\mathcal{M}$) can be evaluated as quickly as possible. It turns out that (slightly modified versions of) time and label elicitation can also be used in this setting. The difference is mainly in the presentation of the pair to be elicited next in the $k$th elicitation step: Instead of presenting it purely randomly, one now applies a prediction heuristic based on the specific choice function under consideration and several previously elicited preference systems of similar users. Specifically, we look at two different prediction heuristics: The simple heuristics of \textit{presenting
pairs of consequences different acts yield under the same state} \citep[Example 1]{JANSEN202269} as well as a more involved heuristic based on the technique of \textit{subgroup discovery} \citep[Example 2]{JANSEN202269}. A graphical illustration of an specific elicitation process based on label elicitation and the first heuristic for a simple decision problem is given Figure~\ref{example_eli}. (For an introduction to subgroup discovery, see, e.g.,~\citet{Ventura2018}.)
\begin{mybox}
\begin{center}
    {\bfseries{Contribution 2}}
\end{center}
Jean Baccelli, Georg Schollmeyer, and Christoph Jansen (2022): Risk aversion over finite domains. \textit{Theory and Decision}, 93: 371–397.\\

\begin{center}
    {\bfseries{Original Abstract}}\\
\end{center}

\textit{We investigate risk attitudes when the underlying domain of payoffs is finite and the payoffs are, in general, not numerical. In such cases, the traditional notions of absolute risk attitudes, that are designed for convex domains of numerical payoffs, are not applicable. We introduce comparative notions of weak and strong risk attitudes that remain applicable. We examine how they are characterized within the rank-dependent utility model, thus including expected utility as a special case. In particular, we characterize strong comparative risk aversion under rank-dependent utility. This is our main result. From this and other findings, we draw two novel conclusions. First, under expected utility, weak and strong comparative risk aversion are characterized by the same condition over finite domains. By contrast, such is not the case under non-expected utility. Second, under expected utility, weak (respectively: strong) comparative risk aversion is characterized by the same condition when the utility functions have finite range and when they have convex range (alternatively, when the payoffs are numerical and their domain is finite or convex, respectively). By contrast, such is not the case under non-expected utility. Thus, considering comparative risk aversion over finite domains leads to a better understanding of the divide between expected and non-expected utility, more generally, the structural properties of the main models of decision-making under risk.}
\end{mybox}
\bigskip
We now turn to the second contribution of Part A. In it, we discuss notions of \textit{comparative risk aversion}, i.e., we seek to compare agents with preference relations $\succcurlyeq_1$ and $\succcurlyeq_2$ on the same domain of von Neumann-Morgenstern lotteries (i.e., finitely supported probability functions) concerning their attitude towards risk. Througout, we assume the agents to be compared, call them Agents 1 and 2, are \textit{ordinally equivalent} in the sense that their preference orders coincide when restricted to the degenerated lotteries (i.e., the lotteries placing all probability mass on one single outcome). The special feature of our setting is that we always assume that the lotteries under consideration are defined on a \textit{finite domain} and thus elude most classical concepts of comparative risk aversion.  \\

Classically, comparative risk definitions often use lotteries defined on \textit{cardinally scaled} spaces as a model primitive. Comparative risk aversion is then often based on the concept of \textit{mean-preserving spreads} (\cite{ROTHSCHILD1970225}), i.e., lotteries coinciding in their expectation, where the riskier one can be obtained by the less risky one by transporting probability mass center outwards \citep[p. 376]{basj2022}. In this case, one can then use the following notion of comparative risk aversion: Agent 1 is (strongly) more risk averse than Agent 2, if, whenever lottery $l_1$ is a mean-preserving spread of lottery $l_2$ and $l_1 \succcurlyeq_1 l_2$, then it necessarily also holds $l_1 \succcurlyeq_2 l_2$. In other words, if Agent 2 mimics every risky decision Agent 1 takes, then 1 is (strongly) more risk averse than 2.\\

The big challenge in our context is that for finite (and potentially non-numeric) domains, the concept of mean-preserving spreads is not well-defined. An alternative spread definition must therefore be used in our context. Natural candidates here are median-preserving spreads \citep{ALLISON2004505,de201817} or more generally quantile-preserving spreads \citep{MENDELSON1987334,BOMMIER20121614}, which function similarly to the definition of the mean-preserving spread mentioned above, only replacing the expected value in the definition with the corresponding quantile. However, these are obviously highly dependent on the choice of the specific quantile used. To free ourselves from this dependency, we use the concept of \textit{spreads} instead \citep{BOMMIER20121614}. Intuitively speaking, a lottery is a spread of another lottery if it is an $r$-quantile preserving spread for at least one $r \in (0,1)$ (for an exact definition see~\citet[Definition 1]{basj2022}). We can then transfer the above definition of comparative risk aversion to our (less-structured) setting: Agent 1 is (strongly) more risk averse than Agent 2, if, whenever lottery $l_1$ is a \textit{spread} of lottery $l_2$ and $l_1 \succcurlyeq_1 l_2$, then it necessarily also holds $l_1 \succcurlyeq_2 l_2$. In other words, if Agent 2 mimics every risky decision Agent 1 takes, then 1 is (strongly) more risk averse than 2 \citet[Definition 2]{basj2022}).\\  

The paper is now mainly devoted to the investigation of comparative risk aversion in three classical models of decision theory under uncertainty: (1) \textit{expected utility (EU)}, (2) \textit{dual expected utility (DEU)} and (3) \textit{rank-dependent utility (RDU)} (see, e.g.,~\citet[Section 2]{basj2022} for the most important facts on these models). For this investigation, based on the utility function $u$ and the probability weighting function $w$ which are the model primitives of RDU, we start by defining two different indices, namely $I_w^{1,2}$ and $I_u^{1,2}$ (see \citet[p. 379]{basj2022}). The main result of the paper is then a characterization of comparative risk aversion in RDU models by means of exactly these indices. The result (Theorem 1) in the paper states the following: \textit{For two (ordinally equivalent) RDU agents with
 function pairs $(w_1,u_1)$ and $(w_2,u_2)$, respectively, Agent $1$ is strongly more risk averse than Agent $2$ if and only if  $I_w^{1,2} \geq I_u^{1,2}.$}\\

Theorem 1 from the paper, which we have just restated and which gives a characterization of strong comparative risk aversion in RDU models, allows two direct conclusions for the EU and DEU models: First, for ordinally equivalent EU agents, we can compare their strong risk aversion by comparing the \textit{strength of concavity}\footnote{See Section 2 of the paper for a definition.} of their respective utility functions \citep[Corollary 1]{basj2022}. Second, for numerical (but still finite) domains, we can compare strong risk aversion of DEU agents by comparing  the \textit{strength of convexity}\footnote{Again, see Section 2 of the paper for a definition.} Thus, for these two types of models the notion of comparative risk aversion simplifies drastically and, at least in the EU case, boils down to our classic understanding of comparative risk aversion.
\bigskip
\begin{mybox}
\begin{center}
    {\bfseries{Contribution 3}}
\end{center}
Christoph Jansen and Thomas Augustin (2022): Decision making with state-dependent preference systems. In: Ciucci, D.; Couso, I.; Medina, J.; Slezak, D.; Petturiti, D.; Bouchon-Meunier, B.; Yager, R.R. (eds): \textit{Information Processing and Management of Uncertainty in Knowledge-Based Systems.} Communications in Computer and Information Science, 1601: 729–742, Springer.\\

\begin{center}
    {\bfseries{Original Abstract}}\\
\end{center}

\textit{In this paper we present some first ideas for decision making with agents whose preference system may depend on an uncertain state of nature. Our main formal framework here are commonly scalable state-dependent decision systems. After giving a formal definition of those systems, we introduce and discuss two criteria for defining optimality of acts, both of which are direct generalizations of classical decision criteria under risk. Further, we show how our criteria can be naturally extended to imprecise probability models. More precisely, we consider convex and finitely generated credal sets. Afterwards, we propose linear pogramming-based algorithms for evaluating our criteria and show how the complexity of these algorithms can be reduced by approximations based on clustering the preference systems under similar states. Finally, we demonstrate our methods in a toy example.}
\end{mybox}
\bigskip
The third contribution of Part A takes us back to the model of decision making under weakly structured information, which we have recalled in Section~\ref{extremes} of the present thesis. In this contribution, we address a much-recognized problem when defining the components of a decision problem, namely that of \textit{state-dependent} utilities and/or preferences: In real-world situations, an agent's preferences often can not be modeled independently of the true state of nature of the underlying decision problem. Prominent examples in this spirit originate, e.g., in insurance science, where often a policyholder's preferences are modelled dependently on their health status (see, e.g.,~\cite{dedonder}).\footnote{Note that the father of these kind of problems is, of course, Newcomb's paradox, as first published in \citet{Nozick1969} and nicely illustrated in~\cite{gilboa2009theory}. However, it would by far from any real-world to call this a real-world application.} Other examples are problems in portfolio selection, where commonly the agent's attitude towards risky choices (and therefore the also preferences) are understood as dependent on some exogenous environment (see, e.g., \cite{WEI}).\\

In these situations, the agent's preferences are state-dependent in the sense that the agent's rankings among the consequences might vary with the knowledge of the true state of nature. Classic theoretical sources on state-dependent preferences and utility functions are \cite{karni2013decision}~or~\cite{ssk1,ssk2}, but many more exist. \cite{jean} provides a modern reappraisal. Most of these use in the classical framework of Anscombe and Aumann of `preferences over horse lotteries' (see~\cite{anscombe}): Starting from a preference relation on the domain of all horse lotteries, they derive `conditional' preference relations for every state fixed and then say the original relation is state-dependent whenever there exist distinct (non-null) states for which these conditional relations differ (see, e.g.,~\cite{Karni1990}).\\

Contribution 3 chooses a more direct approach to model the notion of state dependence. Instead of using the formalism of horse lotteries, we model preferences directly on a finite consequence set and assume that the information about the uncertainty about the states of nature is externally given and can be described by a credal set in the sense of Definition~\ref{credset}. We then allow the agent to express their preferences by means of a preference system as defined in Definition~\ref{ps} \textit{for each state separately}. The only requirements we place on the preference systems assigned to the different states are that they are consistent (see Definition~\ref{consistency}) and \textit{commonly scalable} (see Definition 5 in Contribution 3), where the latter means that all preference systems share common minimal and maximal elements with respect to their ordinal parts.\\

The theoretical part of the paper then starts by adapting two criteria from decision making under weakly structured information (as, e.g., discussed in \cite{JANSEN201849}) to the setting of (consistent and commonly scalable) state-dependent decision problems. The first criterion (compare Definitions 7 and 9 in Contribution 3) relies on the idea of comparing the worst-case expectations of the acts under consideration. Compared to the known situation, the challenge is in particular a suitable definition of the expected utility of an act. However, such a definition can be found by including in the computation -- for each state separately -- the set of all representations of the preference system assigned to the respective state, instead of using only one (state-independent) set of representations as in the known case (see Definition 8 of Contribution 3). The second criterion is based on a generalization of the GSD-relation recalled in Definition~\ref{GSD}. Again, compared to classical GSD, the set of utility representations of the respective preference system is allowed to vary along the different states of nature of the decision problem under uncertainty (see Definition 10 of Contribution 3).\\

The next part of Contribution 3 is devoted to computation. In particular, in Section 4 of the contribution, we derive two linear optimization based algorithms for determining optimal acts with respect to the decision criteria just recalled. Specifically, Theorem 1 of Contribution 3 shows how the worst-case expectation in a state dependent decision problem can be computed by solving a series of linear programs (one for each extreme point of the underlying credal set), whereas Theorem 2 gives a similar collection of linear programs for checking whether two acts are in relation with respect to the state-dependent version of the GSD-relation mentioned earlier. Since these optimization problems (depending on the complexity of the underlying preference systems and the underlying credal set) sometimes have high computational costs, we discuss possibilities for their approximation in the following Section 4.1. of the article. The main idea is to approximate the discussed basic algorithms by grouping the preference systems under in a certain sense similar states of nature. After observing that this grouped version of the original commonly scalable state-dependent decision problem is again commonly scalable (Proposition 3 of the contribution), we propose two concrete procedures approximation (Section 4.3 of the contribution): \textit{Pattern clustering} relies on the idea of grouping the preference systems that contain a certain predefined preference pattern and respectively replacing those by their intersections. In contrast, \textit{distance-based clustering} relies on the idea of grouping those preference systems whose ordinal relations are close with respect to some suitable distance function. The paper ends with an illustrative toy example demonstrating all concepts discussed in the contribution.
\bigskip
\begin{mybox}
\begin{center}
    {\bfseries{Contribution 4}}
\end{center}
Christoph Jansen, Georg Schollmeyer and Thomas Augustin (2023): Multi-target decision making under conditions of severe uncertainty. In:  Torra, V.; Y. Narukawa (eds): \textit{Modeling Decisions for Artificial Intelligence.} Lecture Notes in Computer Science, 13890: 45-57, Springer.\\

\begin{center}
    {\bfseries{Original Abstract}}\\
\end{center}

\textit{The quality of consequences in a decision making problem under (severe) uncertainty must often be compared among different targets (goals, objectives) simultaneously. In addition, the evaluations of a consequence’s performance under the various targets often differ in their scale of measurement, classically being either purely ordinal or perfectly cardinal. In this paper, we transfer recent developments from abstract decision theory with incomplete preferential and probabilistic information to this multi-target setting and show how – by exploiting the (potentially) partial cardinal and partial probabilistic information – more informative orders for comparing decisions can be given than the Pareto order. We discuss some interesting properties of the proposed orders between decision options and show how they can be concretely computed by linear optimization. We conclude the paper by demonstrating our framework in an artificial (but quite real-world) example in the context of comparing algorithms under different performance measures.}
\end{mybox}
\bigskip
I now come to the summary of the last contribution of Part A of the present cumulative habilitation thesis -- Contribution 4. The starting point of the article is, as in Contribution 3, the framework of decision-making under weakly structured information repeated in Section~\ref{extremes}. Specifically, we apply this general framework here to so-called \textit{multi-target decision problems}. In these, the consequences in the set $A$ are not evaluated with one utility function $\phi:A \to \mathbb{R}$, but instead with (finitely many) different utility functions $\phi_1, \dots , \phi_r:A \to \mathbb{R}$, each of which measures the compatibility of a consequence with a certain \textit{target}. As an additional challenge, we assume that not all targets can be interpreted cardinally, but that some of them only have an ordinal scale of measurement (i.e., differences are meaningless apart from their sign).\\

There are many good reasons to model decision situations using multiple targets: In the simplest case, the different targets could compare the short- and long-term benefits of a consequence, as it is relevant in many applications in insurance science (e.g.,~\citet{dedonder}). More complicated cases occur, for example, in benchmark studies for machine learning algorithms: Here, too, it is not uncommon to want to compare the algorithms with regard to several quality criteria (e.g. accuracy and interpretability) at the same time (e.g., \citet{ld2007,ehl2012,de2017joint}). Generally, it seems important to emphasize the difference to the (related) theory of multicriteria decision making (see, e.g.,~\cite{CHAKRABORTY2023100232}~for a survey): While -- roughly speaking -- in multi-criteria decision making the \textit{same utility function} is evaluated with respect to \textit{different criteria}, in the multi-target setting \textit{different utility functions} are evaluated under the \textit{same criterion}. \\

Technically speaking, we proceed as follows in Contribution 4: First, we use the individual targets to form a multi-dimensional target $\phi:=(\phi_1 , \dots ,\phi_r):A \to \mathbb{R}^r$. The range $\phi(A)$ of this multidimensional target is then embedded in a special preference system
\begin{equation}\label{cps}
\text{pref}(\mathbb{R}^r):=[\mathbb{R}^r,R_1^*,R_2^*]   
\end{equation}
(see Equation 4 in the contribution for an exact definition). The system $\text{pref}(\mathbb{R}^r)$ has the following intuitive interpretation: $R_1^*$ can directly be viewed as a componentwise dominance decision,i.e., a consequence is set more desirable than another one if it achieves at least as high value with respect to every target under consideration. In contrast, the construction of the relation $R_2^*$ is slightly more involved: One pair of consequences is preferred to another such pair if it is ensured in the ordinal dimensions that the exchange associated with the first pair is not a deterioration to the exchange associated with the second pair and, in addition, there is component-wise dominance of the differences of the cardinal dimensions.\\

In this setting, the paper proposes to compare acts in multi-target decision problems with respect to a regularized version of the GSD-relation from Definition~\ref{GSD} (see Definition 5 and 10 of Contribution 4 for details). The idea of regularization is inspired by the concepts of \textit{granularity} introduced in \citet[Definition 3]{JANSEN201849}: The expectation dominance in the definition of GSD is no longer demanded for all utility representations of the respective preference system, but only for those that distiguish strict preference over some predefined threshold (more information on this regularization technique in the light of statistical analysis can be also found in Parts B and C of the present habilitation thesis.). Based on this regularized GSD-relation, the paper then goes on defining two (related but) different choice function for multi-target decision problems (see Definition 6 of Contribution 4): The function $\textsf{max}$ selects all acts that dominate all competitors with respect to generalized GSD. Opposed to this, the function $\textsf{und}$ that are not strictly dominated with respect to regularized GSD by any of the competing acts. Note that, by construction, while of course depending on the quality of information, the function will tend to produce large $\textsf{und}$ choice sets, whereas the function $\textsf{max}$ will tend to produce small (or sometimes even empty) choice sets (compare also Proposition 1 of the contribution for some theoretical connections).\\

Theorem 1 of the contribution then proposes an linear-programming based algorithm for evaluating the choice function just recalled. As seen before, the main idea of this algorithm is to decompose the task of determining the choice set of the respective choice functions into several linear programs, roghly one for each extreme point of the underlying credal set. After establishing some theoretical connections between the choice sets based on the regularized GSD-relation for different regularization strenghts (Propostion 3 of the contribution), Section 4 then is devoted to a synthetical application example: Here, we demonstrate the choice functions proposed in the paper in an example with mixed-scaled targets in the context of algortihms comparison.
\subsection{Future Research on Part A}
I will now briefly elaborate on some promising perspectives for future research that (more or less) directly connect to one or more of the articles contributing to Part A of the present thesis. Note that there are plenty of possibilities to follow-up on ideas presented in these papers and that I do not have any ambition to give a complete list of these possibilities here. Let us start with \textit{Contribution 1}. I identify two main avenues for future research here: (I) Theoretical improvement of the elicitation procedures, and (II) real-world applications of the concepts presented. For achieving (I), I can think of (at least) two different directions. Recall that (independent of the concrete procedure) we discussed heuristics for presenting the most promising pair of consequences in each step of the elicitation procedure. On the one hand, as a kind of structural (or \textit{static}) property of the preference system, we have utilized the transitivity of the relation $R_1$. This has allowed us to form the \textit{transitive hull} of the previously elicited pair comparisons after each elicitation step, thus avoiding the retrieval of redundant information that already follows from transitivity. On the other hand, as a form of \textit{dynamic} rule, we have also included data in the form of elicited preference systems of previous users in the prediction. Concretely, we have applied the technique of subgroup discovery here and managed to drastically reduce the number of queries compared to a random presentation in our study.\\

On this basis, a complete system for \textit{rule-based information discovery} could now be built. In a first step, the assumption of externally given uncertainty information (in form of a credal set $\mathcal{M}$ in the sense of Definition~\ref{credset} of this thesis)  will be retained. However -- instead of specific ad hoc rules -- a general framework for avoiding redundant queries under a known \textit{set of rules} shall be achieved. The transitive closure discussed earlier or the prediction of promising queries based on subgroup discovery would then only be prototypical examples of such rules. Further rules could arise, for example, via prior knowledge of patterns in the preference systems under consideration (e.g., top-$k$ orders or interval orders). Overall, this abstract framework will distinguish between \textit{static} and \textit{adaptive} rules, the former being derived from structural properties of the preference system under consideration and the latter being successively determined from data of similar users. Finally, in a second step, capturing the underlying uncertainty will also be included in the query design. In this way, users should be able to make optimal decisions under complex uncertainty by specifying simple (adaptive as well as static) rules and by answering fewest possible simple queries.\\

For achieving (II), i.e.~for finding real-world applications for the elicitation procedures proposed in Contribution 1, several very promising possibilities occur: For instance, the time elicitation method lends itself perfectly naturally to online surveys. Here it would be technically easy to memorize the time from calling up a question to answering it. These times (or more precisely, their reciprocals) could then be used as a proxy for preference intensity, as described in Contribution 1.\footnote{Note such an approach is perfectly in-line with the paradata approach, see, e.g., \citet{Kreuter:2013}.} This could be implemented in recommender systems, for example, which could arrive at a recommendation much faster using the additional implicitly recorded intensity information. But the label elicitation method is also extremely close to reality: users will often have a vague idea of the intensity of a preference expressed with regard to a pair comparison, but will not be able to quantify it precisely with a (cardinally interpreted) number. In such situations, many users will find it much easier to assign the intensity to one of four levels, for example. With the help of label elicitation, this type of partially cardinal information can still be profitably incorporated into the decision-making process.\\

Another, in my view very promising, future research in the context of Part A can transfer the statistical test framework from e.g. \cite{pmlr-v216-jansen23a} (compare also Part C) back into a purely decision-theoretical context. This would open up the possibility of an \textit{empirical decision theory} under weakly structured information that avoids some common but often problematic assumptions of classic decision theory. But step by step: Most commonly, the analysis of decision problems under uncertainty makes use of (and heavily depends on) idealizing assumptions about about the \textit{describability} of the world. We distinguish between two (closely related) types of such assumptions:\\

\textbf{Closed World Assumption (CWA):} It is assumed that a priori -- in the formalization step of the problem -- an exhaustive, informative, and mutually exclusive list of all states relevant for the problem can be given and that the consequence that an act yields under each of  the states is clearly specifiable.\\

\textbf{Small World Assumption (SWA):} It is assumed that there is a natural granularity in which to write down the states of the world of the decision problem under consideration. Any finer or coarser partition of the states of the world is accepted as unnatural.\\

At first glance, it seems obvious to argue that both assumptions are always trivially satisfiable: In the case of CWA, an artificial state $s^*$ could always be introduced, which occurs exactly when none of the clearly specifiable states occurs. However, the CWA assumption also requires that under each state the consequences of the actions must be clearly specifiable. This is obviously less easily satisfied, which makes CWA a very strong assumption after all. Also in the case of SWA, one could argue that a choice function whose choice sets depend on the concrete framing of the decision problem should not be chosen anyway, and thus invariance under the granularity of the set of states of the world should be considered as a quality measure for choice functions. But the requirement of perfect invariance under granularity quickly leads to trivial decision criteria and is therefore presumably only desirable at first sight. It is also important to note here that the two assumptions are by no means independent of each other: if artificial state $s^*$ is introduced as described above, it is probably only in extremely rare cases possible to make statements about the naturalness of its granularity.\\ 

Overall, it can be said that the basic model of decision theory under uncertainty depends on strong assumptions that are difficult to meet. This considerably limits its applicability and thus also its practical usefulness in certain areas. In fact, radically different approaches to modeling decision situations have been proposed to avoid it (e.g.,~\cite{gilboa1995case,s2016,BLUME2021105306}). On the contrary, M
my planned path for future research aims at retaining the appeal and simplicity of the original theory but completely avoids the need to specify the states of the world. In this way, assumptions such as CWA and OWA are also avoided, which -- hopefully -- can help decision theory to find further fields of application. \\

The main idea of the proposal sketched here is to address decision problems in a radically empirical way: instead of specifying states and consequences a priori to the decision analysis, we assume our decision problem to consist only of \textit{observable} components of the form \textit{act $\rightarrow$ consequence}, i.e.~of a protocol of act-consequence-pairs storing the information which act lead to which consequence in previous trials of choosing the corresponding act. Based on these protocols one could then go on define an \textit{empirical choice function}, which are designed as empirical analogs of the respective choice function one would deem appropriate for the theoretical decision problem underlying. To be a bit more precise, as before, denote by $\mathcal{G}$ the finite set of acts of the form $X:S \to A$ of the underlying theoretical decision problem (still assuming $A$ is ordered by a preference system). Moreover, assume we have available a protocol $\mathscr{P}_n$ consisting of $n\cdot |\mathcal{G}|$ act-consequence-pairs as described above, $n$ for each act in $\mathcal{G}$.\footnote{Note that the assumption of equally many act-consequence pairs for each act is only for sake of notational simplicity. Nothing hinges on that.} Finally, for every $\mathcal{H} \subseteq \mathcal{G}$, denote by $S_n(\mathcal{H})$ that subset of $\mathscr{P}_n$ containing only those act-consequence pairs with acts lying in $\mathcal{H}$, and set $S(\mathscr{P}_n):=\{S_n(\mathcal{H}): \mathcal{H} \subseteq \mathcal{G}\}$. \\

An \textit{empirical choice function (ecf)} $\hat{\text{ch}}_n: S(\mathscr{P}_n) \to 2^{\mathcal{G}}$ is then simply a function satisfying $\hat{\text{ch}}_n(S_n(\mathcal{H})) \subseteq \mathcal{H}$ for all $\mathcal{H} \subseteq \mathcal{G}$, i.e., a function assigning sub-protocols of act-consequence pairs to subsets of the acts which are relevant in these sub-protocols.  Two observations in the context of ecfs come to mind: First, it is indeed not necessary to specify the state space of the underlying decision problem for applying an ecf. Thus, strong assumptions such as OWA and CWA are completely avoided. Secondly, however, it is initially questionable how the choice sets of an empirical ecf relate to those of the true choice function, which the ecf is supposed to approximate. Consequently, desiderata are necessary that are placed on the relationship between the two functions. While an exhaustive list of such desiderata would certainly overstretch the provisional character of this presentation, one very natural candidate can already be mentioned here. For that, assume ch$:2^{\mathcal{G}} \to \mathcal{G}$ the choice function to be approximated by $\hat{\text{ch}}_n: S(\mathscr{P}_n) \to 2^{\mathcal{G}}$. Then, the following may be demanded: \\

\textbf{Consistency:} If $\pi$ is a probability measure on the states $S$ and, for every $X \in\mathcal{G}$, the pairs in $S_n(\{X\})$ form an \textit{i.i.d.}-sample from $\pi_{X}$, then
 \begin{equation}
    \forall \mathcal{H} \subseteq \mathcal{G}:~
        \pi \Biggl(\Bigr\{s \in S: \lim_{n\to \infty} \hat{\text{ch}}_n(S_n(\mathcal{H}))=\text{ch}(\mathcal{H}) \Bigl\} \Biggr)=1,
    \end{equation}
 where convergence is defined via the trivial metric assigning $0$ to identical pairs of sets, and $1$ to non-identical.\\

In other words, a very natural desideratum to connect an ecf to the choice function it intends to approximate from the protocol $\mathscr{P}_n$ would be to demand that the choice sets it produces are strongly consistent statistical estimators for the true choice sets, assuming \textit{i.i.d.}-sampling design. First steps towards further exploring the ideas just outlined would be to first find meaningful ecfs for the most common choice functions in decision theory, then investigate which of these are consistent in the above sense, and finally thinking about desiderata for ecfs beyond consistency. A look at Contribution 8 (Theorem 3.6) gives hope for the feasibility of these steps: In a slightly more specific setting and under mild regularity conditions, we there show that the empirical version of the choice function induced by the GSD-relation is a strongly consistent estimator for the underlying population variant. This result seems to straightforwardly generalize to the considerations just sketched. 
\newpage
\section{Machine Learning under Weakly Structured Information}
\noindent 
\\
\begin{minipage}{0.3\textwidth}

\end{minipage}
\hfill
\begin{minipage}{0.5\textwidth}
\textit{A decision-theoretic grounding promises to support the systematic development of prescriptive ML methods
in a principled (rather than ad hoc) manner.}\\[.2cm] \hfill --------------------------------------(\citet[p. 5]{hüllermeier2021prescriptive})
\end{minipage}\\[.5cm]
I now come to the second main part of this habilitation thesis: the application of decision theory under weakly structured information to machine learning problems. I will start by carefully demonstrating how the decision-theoretic concepts recalled in Part A of this thesis can be transferred to machine learning under weakly structured information, e.g. under non-standard data or non-typical scales of measurement. Afterwards, I will give brief general summaries of the four contribution building the core of Part B. Finally, I elaborate on promising avenues for future research projects based on the work presented.
\subsection{A Decision-Theoretic Perspective on Machine Learning}
\subsubsection{Preference Systems Revisited} \label{revi}
We now want to address the question of how the concept of the preference system from Definition~\ref{ps}, which is motivated purely by decision theory, can be transferred to problems from statistics and machine learning. Let us first recall the actual interpretation of a preference system. Let $\mathcal{A}=[A, R_1,R_2]$ be an (arbitrary) preference system. In the context of decision theory, the relation $R_1$ models the classical desirability relation: If it is true for consequences $a,b \in A$ that $(a,b) \in R_1$, then we interpret this as \textit{$a$ is at least as desirable as $b$}. Moreover, if for consequences $a,b,c,d \in A$ it holds that $((a,b),(c,d)) \in R_2$, this means that\textit{ exchanging $b$ for $a$ is at least as desirable as exchanging $d$ for $c$.} In other words, for the latter, we can imagine that we have objects $b$ and $d$ and we are offered the option of swapping either $b$ for $a$ or $d$ for $c$ (but not both) and that in this case we opt for the former.\\

But what do these - initially purely decision-theoretical considerations - have to do with problems of statistics and machine learning? Can they perhaps even be used to contribute useful new concepts here as well? The answer to this is clearly yes! It turns out that preference systems can very naturally be used to describe data (or more precisely sets) that do not have a typical scale of measurement. Before I specify this in more detail, let's repeat (somewhat informally) the classic scale levels for the sake of completeness: Consider some random variable $X:\Omega \to A$ mapping to $A$. We distinguish:
\begin{itemize}
    \item If $A$ is structured by a preorder (but nothing else), we call $A$ of \textit{ordinal scale}. In this case, the set $\mathcal{U}_{all}$ of all candidate scales $u:A \to \mathbb{R}$ that are isotone with respect to this preorder as a whole represents the structural information on $A$. Any analysis of the variable $X$ should be \textit{invariant} under the choice of the candidate scale $u \in \mathcal{U}_{all}$. Observe that, expressed via preference systems, this exactly corresponds to the case where $R_2$ is chosen to be the trivial preorder, consisting only of $\{(a,a): a \in A\}$.
    \item If the order on $A$ is induced by some metric $d$, then we call $A$ of \textit{cardinal scale.}
    In this case, there exists a scale $u^*:A \to \mathbb{R}$ that is unique (up to some neglectable transformation class). Here, any analysis of the variable $X$ should be based on $u^*$ alone. Note that in terms of preference systems this corresponds to the case that the preference system under consideration forms a \textit{positive difference structure} in the sense of~\citet[Chapter 4]{krantz}.
\end{itemize}
Preference systems now give us the possibility of making the two standard scales just repeated appear only as extreme poles of a continuous spectrum of different scales of measurement. In particular, they allow us to model sets whose structure may vary locally: While some areas of the considered set are highly ordered and perhaps even provided with a corresponding distance concept, other areas of the set may have little or no order structure at all. In this case, we use $R_1$ to model the available ordinal, and $R_2$ to express the available cardinal information. Intuitively, we can then say that
\begin{itemize}
    \item $A$ is \textit{locally almost cardinal} on subsets where $R_1$ and $R_2$ are very dense. Any analysis of the variable $X$ should be based on the set $\mathcal{U}_{\mathcal{A}}$ alone, which reflects this local property by implicitly allowing the compatible candidate scales low variability on the dense parts. 
    \item $A$ is \textit{locally at most ordinal} on subsets where $R_2$ is sparse or even trivial. Any analysis of the variable $X$ should be based on the set $\mathcal{U}_{\mathcal{A}}$ alone, which reflects this local property by implicitly allowing the compatible candidate scales high variability on the sparse parts.
\end{itemize}
\subsubsection{Regularization in Preference Systems} \label{reginpref}
A further transfer from the decision-theoretical to the statistical setting in the context of preference systems succeeds in the area of regularization: While the granularity parameter $\delta$ from Definition~\ref{granularity} served as a kind of lower bound for the noticeability of utility differences in the decision-theoretical context, it can be understood as a regularization parameter in the statistical reading recalled before. The idea of regularization is simply the following: In the context of statistics and machine learning, the set $A$ (and thus also the order structure prevailing here) will often only be empirically accessible. More precisely, instead of the true preference system $\mathcal{A}$, one usually considers the subsystem of $\mathcal{A}$, which is spanned by a concrete sample $X_1 , \dots , X_n$ in $A$. For ease of exposition, let us denote this "empirical" subsystem by $\hat{\mathcal{A}}_n$ for the moment.\\

As usual, in such a case, any statistical analysis is based on the corresponding empirical analogs derived from the sample. In the case of preference systems, this implies that any further analysis should depend only on the set of scales -- $\mathcal{U}_{\hat{\mathcal{A}}_n}$ -- compatible with the empirical preference system. This is where regularization comes into play. If you want to make the underlying empirical preference system less prone to picking up the noise created by the sampling, it makes sense to base the further analysis on the normalized representation set $\mathcal{N}^{\delta}_{\hat{\mathcal{A}}_n}$ instead of $\mathcal{U}_{\hat{\mathcal{A}}_n}$, where $\delta \in (0,1]$ is the regularization strength. In this way, the effect is counteracted by sorting out too extreme utility representations, which only come about due to the specific form of the noise in the sample, with the help of regularization. \\

\begin{figure}[h]
    \centering
    \includegraphics[width=.45\linewidth]{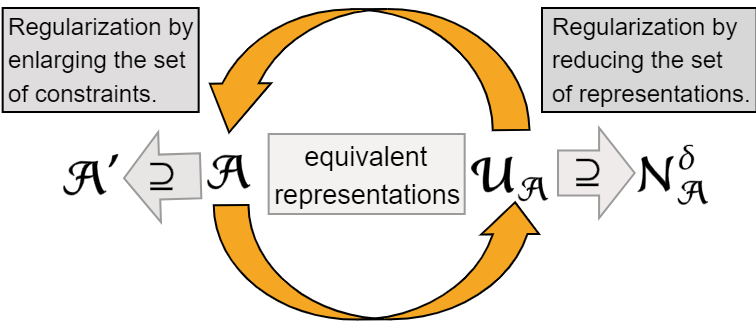}
    \caption{The diagram shows a schematic comparison of the two different types of regularization in preference systems. Note that the identical diagram can be found in \citet[Section 3]{pmlr-v216-jansen23a}.}
    \label{example_regu}
\end{figure}

Since this form of regularization can be controlled by a specific parameter, we call it \textit{parameter-driven} regularization (see \citet[Section 3]{pmlr-v216-jansen23a}). This is in contrast to another form of regularization in preference systems, the so-called \textit{order-theoretic} regularization (again, see \citet[Section 3]{pmlr-v216-jansen23a}): This attempts to mitigate the influence of sampling noise by specifically adding order constraints to the empirical preference system. To ensure that the added ordering constraints are not arbitrary and the resulting regularization is therefore misguided, they must of course be well thought out. This regularization method is therefore particularly suitable when expert knowledge or other global information is available a priori. A schematic comparison of the two types of regularization, which is taken from \citet[Section 3]{pmlr-v216-jansen23a}, can be found in Figure~\ref{example_regu}.\footnote{Note that this distinction is perfectly align with the more common classificaon of regularization-types in Ivanov-, Tikhonov- and Morozov-regularization (see, e.g.,~\citet{ora2016}): Both proposed approaches are special types of Ivanov-regularization.}\\

\subsubsection{The Special Case of Multidimensional Spaces with Differently Scaled Dimensions} \label{mdsdsd}
So far I have talked about the situation of data in general preference systems. Now I would like to turn to an important special class of such data (or sample spaces): \textit{Multidimensional spaces with differently scaled dimensions}. These occur quite naturally whenever a general phenomenon is to be described in terms of several facets, which do not necessarily all have the same scale of measurement. Natural examples within classical statistics can be found, for example, in multivariate poverty measurement, where, at the latest since the capability approach by Sen \citep{sen1985commodities}, there is wide consensus that poverty has more facets than only income or wealth and that also potential ordinal criteria like education are influential here. These statistical examples will be discussed in greater depth in Part C of this habilitation thesis.\\

But also in modern machine learning, very natural examples of this type of sample space can be found quickly. The most obvious of these examples is the benchmarking of algorithms with respect to different quality metrics simultaneously while additionally taking into account the randomness of the benchmark suite. Quality metrics of different scales of measurement often play a role here as a matter of course: In addition to classic cardinally interpretable criteria such as accuracy or the Brier score, ordinal criteria such as, e.g., feature sparseness as a proxy for interpretability \citep{schneider2023multi}, risk levels in the EU AI act \citep{laux2024trustworthy} or other regulatory frameworks \cite{schmitt2022mapping}, also play a crucial role here.\\

But how can preference systems actually be used to model such multidimensional spaces with differently scaled dimensions in an information-exhaustive way? An answer to this question requires some additional notation. Concretely, we address $r \in \mathbb{N}$ dimensional spaces for which we assume --  w.l.o.g.~-- that the first $0 \leq z \leq r$ dimensions are of cardinal scale (implying that differences of elements may be interpreted as such), while the remaining ones are purely ordinal (implying differences to be meaningless apart from the sign). Interestingly, this \textit{partial cardinal structure} can be formalized perfectly natural by considering (bounded subsystems of) the preference system
\begin{equation}\label{cps}
\text{pref}(\mathbb{R}^r)=[ \mathbb{R}^r,R_1^*,R_2^*]   
\end{equation}
where:
\begin{equation} \label{cps1}
 R_1^*=\Bigl\{(x,y): x_j \geq y_j ~\forall j\leq r \Bigr\}
 \end{equation}
 \begin{equation}  \label{cps2}
 R_2^*  =  \Biggl\{((x,y),(x',y')): \begin{array}{lr}
        x_j -y_j \geq x_j'-y_j' ~~\forall j\leq z\\
        x_j \geq x_j'\geq y_j' \geq  y_j ~\forall j> z
    \end{array}\Biggr\}
\end{equation}
The partial order $R_1^*$ can be interpreted as a simple component-wise dominance relation and, thus, captures the entire ordinal information encoded in the data. In contrast, the preorder $R_2^*$ captures the entire information in the cardinal dimensions: One pair of vectors is preferred to another one if it is ensured in the ordinal dimensions that the exchange associated with the first pair is not a deterioration to the exchange associated with the second pair and, in addition, there is component-wise dominance of the differences of the cardinal dimensions. In perfect accordance with the considerations in Section~\ref{revi}, any analysis of this special type of non-standard data should then be based on the set $\mathcal{U}_{\text{pref}(\mathbb{R}^r)}$ of utility function representing the associated preference system alone.
\subsubsection{The Role of Generalized Stochastic Dominance}
We now come to a further transfer of an originally genuinely decision-theoretical concept into the field of statistics and machine learning, namely that of generalized stochastic dominance. Having previously reinterpreted preference systems in this context, this further transfer is now an easy one: We simply transfer the definition of the GSD relation as we got to know it in Definition~\ref{GSD} to the new interpretation of the preference systems used. The main difference here lies in the interpretation: While GSD was previously considered between actions in a decision problem under uncertainty and accordingly regarded as the basis of a choice function, it is now understood as a generalized expectation ordering between random variables in a sample space with non-standard scale of measurement. As can be seen in the four publications of Part B summarized below, this transfer is also extremely fruitful and provides many new insights that would not be possible without a decision-theoretical perspective.
\subsection{Contributions to Part B}
I will now briefly summarize the four publications forming the core of Part B of the present work. The order in which the publications are discussed is chronological by date of publication, starting with:
\bigskip
\begin{mybox}
\begin{center}
    {\bfseries{Contribution 5}}
\end{center}
Christoph Jansen, Malte Nalenz, Georg Schollmeyer, and Thomas Augustin (2023): Statistical comparisons of classifiers by
generalized stochastic dominance. \textit{Journal of Machine Learning Research}, 24(231): 1–37.\\

\begin{center}
    {\bfseries{Original Abstract}}\\
\end{center}

\textit{Although being a crucial question for the development of machine learning algorithms, there is still no consensus on how to compare classifiers over multiple data sets with respect to several criteria. Every comparison framework is confronted with (at least) three fundamental challenges: the multiplicity of quality criteria, the multiplicity of data sets and the randomness of the selection of data sets. In this paper, we add a fresh view to the vivid debate by adopting recent developments in decision theory. Based on so-called preference systems, our framework ranks classifiers by a generalized concept of stochastic dominance, which powerfully circumvents the cumbersome, and often even self-contradictory, reliance on aggregates. Moreover, we show that generalized stochastic dominance can be operationalized by solving easy-to-handle linear programs and moreover statistically tested employing an adapted two-sample observation-randomization test. This yields indeed a powerful framework for the statistical comparison of classifiers over multiple data sets with respect to multiple quality criteria simultaneously. We illustrate and investigate our framework in a simulation study and with a set of standard benchmark data sets.}
\end{mybox}
\bigskip
I now come to the summary of the first article of Part B of this cumulative habilitation thesis. In it, we offer a new perspective on a much-discussed problem in machine learning: the systematic benchmarking of classifiers using multiple quality metrics simultaneously, compared across different datasets. To enable a systematic study, we first identify three different levels of challenges when comparing classifiers:
\\[.15cm]
\textit{Level 1:} In the case of multiple quality metrics, two classifiers can generally not be trivially compared already on \textit{one single} data set. Without further assumptions, no decision between the classifiers can be made in situations of conflicting metrics: The component-wise dominance relation is only a \textit{partial order}. Moreover, the intensity information encoded in the cardinal quality metrics remains unused.
\\[.15cm]
\textit{Level 2:} Even if the problem in Level~1 can be circumvented (e.g., if there indeed happens to hold component-wise dominance), the order among classifiers that holds over one fixed data set may change or even completely reverse over another data set. This makes the comparison of classifiers a decision problem \textit{under uncertainty} about the data sets. This uncertainty should be adequately included in any further analysis.
\\[.15cm]
\textit{Level 3:} Since both the set of all relevant data sets and their probability distribution will in general be unknown, instead of the decision problem from Level 2, it will often be necessary to analyze an empirical analog of the problem over a sample of data sets. This means that even if the problems of Levels 1 and 2 can be circumvented and a meaningful order of classifiers for the concrete sample of data sets can be found, a different order of classifiers could occur as soon as another sample of data sets is considered. The solution of such an empirical decision problem is subject to \textit{statistical uncertainty}. It is desirable to be able to control this statistical uncertainty by constructing a suitable statistical test. \\

Contribution 5 aims to address all three problem levels simultaneously, proposing a general framework that can fully utilize the information in the mixed-scale performance metrics. While much attention has already been paid to approaches to meet the challenges of Levels 2 and 3 in the case of a \textit{single} quality metric (see, e.g., \citet{demvsar2006statistical,Garca2008AnEO,GARCIA20102044}) for frequentist or, e.g.,~\citet{benavoli2016should,corani2017statistical,bcdz2017} for Bayesian approaches), and approaches already exist that address Levels 1 and 2 (see, e.g., \citet{ehl2012,de2017joint}), to the best of our knowledge we have searched in vain for holistic approaches accounting for all three levels simulatneously.\\ 

To illustrate how exactly our methodological framework addresses the problems of the respective level, we first need a little additional notation. Let 
 \begin{itemize}
     \item $\mathcal{D}$ denote the universe of \textit{data sets} that are relevant for the classification task in question,
     \item $\mathcal{C}$ denote a finite set of \textit{classifiers} applicable on the data sets from $\mathcal{D}$, 
     \item $\phi_i: \mathcal{C} \times \mathcal{D} \to \mathbb{R}$ denote a \textit{quality metric} for every $i \in \{1, \dots , n\}$ (higher values are better),
     \item $\phi:=(\phi_1 , \dots , \phi_r):  \mathcal{D} \times \mathcal{C} \to \mathbb{R}^r$ the composed \textit{multidimensional} quality metric.
 \end{itemize}
For a data set $D \in \mathcal{D}$ and a classifier $C \in \mathcal{C}$, the \textit{reward} $\phi_i(C,D)$ is then interpreted as the quality of the classifier $C$ for data set $D$ with respect to the quality metric $\phi_i$. Importantly, note that the different quality metrics $\phi_1 , \dots , \phi_r$  \textit{are not} assumed to be of the same scale of measurement. Instead, we assume that  $\phi_1 , \dots , \phi_z$, $z \leq r$, are of cardinal scale (implying that differences of elements may be interpreted as such), while the remaining ones are purely ordinal (implying differences to be meaningless apart from the sign).
\\

The key idea is now to embed the image $\phi(\mathcal{C} \times \mathcal{D})$ of the multidimensional quality metric $\phi$ in the preference system $\text{pref}(\mathbb{R}^r)=[ \mathbb{R}^r,R_1^*,R_2^*] $ defined in Equations (\ref{cps}), (\ref{cps1}), and (\ref{cps2}) of the present thesis. As described there, this allows us to ideally exploit the partial cardinal information encoded in the mixed-scaled performance metrics, with neither having to rely purely on the component-wise partial ordering nor making overly optimistic or unjustified assumptions about the weighting of the different metrics. Based on this embedding, we then propose to compare classifiers by applying a regularized version of the GSD-relation recalled in Definition~\ref{GSD} of the present thesis (the formal details of this regularized GSD-relation among classifiers can be found in Definition 6 of Contribution 5).\\

Roughly, stated this relation looks as follows: If $\pi$ is the probability measure gernerating the data sets from $\mathcal{D}$, and $C,C' \in \mathcal{C}$ are two distinct classifiers, and $\delta \in [0,1]$ is some suitable\footnote{I refer the reader to Contribution 5 for the technical details.} number, then we say that $C$ $\delta$-dominates $C'$, denoted by $C \succsim_{\delta} C'$, 
if $$\mathbb{E}_{\pi}(u \circ \phi(C,\cdot)) \geq \mathbb{E}_{\pi}(u \circ \phi(C',\cdot))$$ 
for all normalized representations $u \in \mathcal{N}^{\delta}_{\text{pref}(\mathbb{R}^r)}$ respecting the threshold $\delta$. Note that for $\delta =0$ this (essentially) coincides with the random variable $\phi(C,\cdot)$ dominating the random variable $\phi(C', \cdot)$ with respect to the generalized stochastic dominance relation $R_{(\text{pref}(\mathbb{R}^r),\{\pi\})}$ as recalled in Definition~\ref{GSD} of the present thesis.\\

Comparing classifiers by the relation $\succsim_{\delta}$ as just recalled, we can indeed address Levels 1, 2, and 3 from before within the same formal framework: First of all, by working with the set $\mathcal{N}^{\delta}_{\text{pref}(\mathbb{R}^r)}$ of all representations of the embedded preference system, we ensure that all (partial) ordering information -- ordinal and cardinal -- is fully reflected in the comparison of the classifiers. This explicitly addresses the challenge from \textit{Level 1}, as we are neither restricted to working only with the component-wise partial order (wasting the partial cardinal information) nor have to make overly optimistic assumptions about the weighting of the different quality metrics (ignoring that our cardinal information is only partial). Furthermore, the parameter $\delta$ allows us to increase the ordering power of our relation if necessary by weakening the influence of all quality metrics involved to the same extent. The challenges of \textit{Level 2} are addressed by including the distribution $\pi$ of the data sets: By choosing a criterion based on (generalized) expected performances, it is ensured that all possible (and not just a selection of) data sets are included (at least on a theoretical level). In this way, the problem of comparing classifiers is treated as the decision problem under uncertainty that it actually is. The most difficult challenges are undoubtedly those of \textit{Level 3}: In real-world problems, we have neither the set $\mathcal{D}$ nor the probability measure $\pi$ available and therefore the relation $\succsim_{\delta}$ can not be evaluated. Consequently, as usual in statistics and machine learning, we have to deal with empirical analogs based on a sample of data sets. However, to be able to control the resulting statistical uncertainty (or, more precisely, the type-I error), we need an adequate statistical test. However, since we are in a completely non-parametric setting, we use a non-parametric, or more precisely, a permutation test.\\

The main findings of the paper can now be summarized as follows: After transferring the formal framework of GSD to the problem of multivariate classification comparison in Section 3 and looking at some interesting properties of the relation $\succsim_{\delta}$ (compare Propositions 1 and 2 of Contribution 5), we in detail describe the permutation test used in Section 4. To this end, we first derive a suitable test statistic and show how it can be calculated efficiently (see Proposition 3 of Contribution 5). We then describe a step-by-step test scheme and discuss how this can be reduced to tests known from the literature in special cases. The rest of the paper is then devoted to the application: First, we conduct a simulation study comparing our proposed test with two variants of the test from \citet{demvsar2006statistical} adapted to the multi-criteria setting (see Section 5 of Contribution 5). Afterwards, in Section 6 of Contribution 5, we demonstrate the power of our framework on a suite of benchmark data sets originating from the UCI repository \citep{graf}. Both studies make perfectly clear the advantages of the GSD-based approach to classifier comparison: While the adapted \citet{demvsar2006statistical} test (called \textit{all test} in Contribution 5) turns out to be not very sensitive for the null hypothesis, the contrary seems to hold for the proposed GSD-test. Especially, the regularization of the test statistic by means of the parameter $\delta$ discussed before shows to have a positive influence of the test's sensitivity: Increasing the regularization strength empirically shows (in the simulation study) to improve the sensitivity of the test statistic. Finally, the inclusion of partial cardinal information via $R_2^*$ also has a clear effect: In the case of classical stochastic dominance, no significant correlation can be found (Section 6.3 of Contribution 5).
\bigskip
\begin{mybox}
\begin{center}
    {\bfseries{Contribution 6}}
\end{center}
 Hannah Blocher, Georg Schollmeyer, Malte Nalenz and Christoph Jansen (2024): Comparing Machine Learning Algorithms by Union-Free Generic Depth. \textit{International Journal of Approximate Reasoning}, 169: 1-23.\\

\begin{center}
    {\bfseries{Original Abstract}}\\
\end{center}

\textit{We propose a framework for descriptively analyzing sets of partial orders based on the concept of depth functions. Despite intensive studies of depth functions in linear and metric spaces, there is very little discussion on depth functions for non-standard data types such as partial orders. We introduce an adaptation of the well-known simplicial depth to the set of all partial orders, the union-free generic (ufg) depth. Moreover, we utilize our ufg depth for a comparison of machine learning algorithms based on multidimensional performance measures. Concretely, we analyze the distribution of different classifier performances over a sample of standard benchmark data sets. Our results promisingly demonstrate that our approach differs substantially from existing benchmarking approaches and, therefore, adds a new perspective to the vivid debate on the comparison of classifiers.}
\end{mybox}
\bigskip
Contribution 6 (see \citet{Blocher_isipta, BLOCHER2024109166}) is dedicated to a very similar question as Contribution 5: The comparison of machine learning algorithms with regard to several performance metrics, over an entire sample (or suite) of data sets. The first important difference is that Contribution 6 analyzes purely descriptively. The challenges of Level 3 (see description of Contribution 5) are therefore explicitly ignored. The main difference, however, lies in the methodology: Instead of comparing the algorithms using a GSD-based approach as before, we now use an approach based on \textit{data depth} (see \citet{tukey75,SERFLING2004,serfling2006depth} for original sources). The main tool here are so-called \textit{depth functions}: These assign -- first generally speaking -- a depth value to each data point, which indicates its \textit{centrality} or \textit{outlyingness} with respect to a data cloud (or more generally a probability distribution). \\

While originally used for two-dimensional (real-valued) data sets, the theory of data depth was in the sequel also adapted to $\mathbb{R}^d$-valued data (see, e.g., (\citet{liu99,SERFLING2004,wang2005}). These extensions then enabled a completely new possibility of robust data analysis, as they allow natural correspondences of the concept of quantiles also for multivariate data structures and thus in particular induce the identification of outliers and central points within multivariate data clouds (or more generally distributions). Consequently, many different proposals for concrete depth functions on $\mathbb{R}^d$ were subsequently made (e.g.,~\citet{barnet1976ordering,liu90,mosi}) and also axiomatic approach to these were devised (see, e.g., \citet{sz2000}). Moreover, also generalizations of depth functions to abstract metric spaces (e.g.,~\citet{goibert2022statistical,geenens2023statistical}) or lattice-valued data (e.g.,~\citet{schollmeyer17a}) attracted a lot of attention.
  \\

Our data depth-based comparison framework for machine learning algorithms now goes exactly in the latter direction: We propose a depth function defined on the set of all \textit{partial} orders.\footnote{As this often leads to confusion: The proposed depth function lives on the set of all partial orders (and not on an arbitrary partially ordered set). In particular, each partial order within this set is assigned a depth value that measures how central it is in the set of all partial orders (or in a specific sample of partial orders). In the context of data analysis, each data point is then a partial order (as a generalization of e.g. ranking data) and not just a value in a partially ordered set.} This depth function, which bears the name \textit{union-free generic depth (ufg depth)} (see Section 3 of Contribution 6 for an exact definition), can be understood as a direct generalization of the simplicial depth \citep{liu90} on $\mathbb{R}^d$ to this (non-standard) data type. Roughly speaking, the (empirical) simplicial depth of a real vector is the probability that this vector (identified as a point in $\mathbb{R}^d$) lies in the triangle spanned by three randomly drawn points from the underlying data cloud. As we lack structure when trying to apply this exact definition to the set of all partial orders, we need a modified version that is still well-defined here. It turns out that this is indeed possible: We replace the set of all triangles, as used in classical simplicial depth, by a special set system over the set of all partial orders (called $\mathscr{S}$, see Section 4 of Contribution 6). This set system consists of all sets of partial orders that are \textit{union-free} (see Equation (C2) of the contribution) on the one hand and \textit{generic} (see Equation (C1) of the contribution) on the other, where these are two properties that abstract the idea of triangles in the simplicity depth to our setting.\\

Equipped with this depth function on the set all partial orders, we can now turn to the original goal, namely the comparison of machine learning algorithms with regard to several performance metrics, over an entire sample (or suite) of data sets. For this, first observe that under a multidimensional performance metric $\phi=(\phi_1 , \dots , \phi_r)$ (compare the discussions concerning Contribution 5 in the present thesis), and a finite set $\mathcal{C}$ of algorithms (again compare the discussion concerning Contribution 5), each data set produces a partial order on the set $\mathcal{C}$: We set classifier $C$ at least as good as $C'$ on data set $D$, whenever $\phi_i(C,D) \geq \phi_i(C',D)$ for all $i \leq r$. From this point of view, the classical task of\textit{ benchmarking algorithms} with respect to several performance metrics provides a \textit{random sample} in the set of all \textit{partial orders}.\\

This is exactly where our application comes into play: Given a sample of partial orderings (based on multidimensional performance comparisons), we want to analyze the centrality and outlyingness, roughly speaking the \textit{typicality}, of partial orderings over the algorithms under investigation. We use the suggested ufg depth for this. The main contributions of the paper can now be summarized as follows: In Section 4 of Contribution 6, we give a formal definition of the ufg depth (see Definition 3 for the population and Definition 4 for the empirical variant) and all concepts which it is based upon. Section 5 is then devoted to an in-depth investigation of some theoretical properties of the ufg depth. The section is split in two parts. We start by discussing some properties of the set system $\mathscr{S}$ mentioned above in Section 5.1: While Theorem 3 delivers fundamental insights in the types of sets belonging to $\mathscr{S}$, Theorems 4 and 5 give lower and upper bounds for the cardinality of elements of $\mathscr{S}$. Section 5.2 then discusses properties of the ufg depth itself: While Theorem 7 states that the ufg-depth cannot be broken down to sum-statistics and, therefore, is indeed dependent on the inserted partial order as a whole, Theorem 10 establishes consistency of the empirical ufg depth in the sense that the empirical ufg depth converges almost surely to its associated population variant.\\

The rest of the paper is then concerned with applications on classifier comparison: We apply our methodology to two different standard benchmark suites, namely UCI \citep{graf} and OpenML \citep{OpenML2013}. Moreover, we compare our methodology with other ones, also aiming at benchmarking classifiers with respect to several performance metrics simultaneously, specifically the framework of \citet{JMLR} (Contribution 5) as well as the adapted Bradley-Terry-Luce framework discussed in \citet{davidson70}. The analyses show very interesting results that are too extensive to summarize here. The reader is referred directly to the corresponding sections in Contribution 6. However, it is important to emphasize that our framework, far beyond the ones used for comparison, does not only provide a specific order of classifiers, but instead allows to analyze the \textit{distribution} of the partial orders and, therefore, make statements about their typicality. This gives a unique appeal to the framework presented here.
\bigskip
\begin{mybox}
\begin{center}
    {\bfseries{Contribution 7}}
\end{center}
Julian Rodemann, Christoph Jansen, Georg Schollmeyer and Thomas Augustin (2023):  In All Likelihoods: Robust Selection of Pseudo-Labeled Data. In: Miranda, E.; Montes, I.; Quaeghebeur, E.; Vantaggi, B. (eds): \textit{Proceedings of the Thirteenth International Symposium on Imprecise Probabilities: Theories and Applications.} Proceedings of Machine Learning Research, 215: 421-425, PMLR.\\

\begin{center}
    {\bfseries{Original Abstract}}\\
\end{center}

\textit{Self-training is a simple yet effective method within semi-supervised learning. Self-training’s rationale is to iteratively enhance training data by adding pseudo-labeled data. Its generalization performance heavily depends on the selection of these pseudo-labeled data (PLS). In this paper, we render PLS more robust towards the involved modeling assumptions. To this end, we treat PLS as a decision problem, which allows us to introduce a generalized utility function. The idea is to select pseudo-labeled data that maximize a multi-objective utility function. We demonstrate that the latter can be constructed to account for different sources of uncertainty and explore three examples: model selection, accumulation of errors and covariate shift. In the absence of second-order information on such uncertainties, we furthermore consider the generic approach of the generalized Bayesian $\alpha$-cut updating rule for credal sets. We spotlight the application of three of our robust extensions on both simulated and three real-world data sets. In a benchmarking study, we compare these extensions to traditional PLS methods. Results suggest that robustness with regard to model choice can lead to substantial accuracy gains.}
\end{mybox}
\bigskip
I now turn to summarzing the third contribution to Part B of the present cumulative habilitation thesis: Contribution~7. This article covers another aspect of machine learning under weakly structured information, namely the robustification of \textit{pseudo-labeling (PL)} (sometimes also called \textit{self-training}, see, e.g., \citet{triguero2015self}) by using credal sets, multiple models, and weak labeling in the definition of the criterion used for the selection of the pseudo-labeled data. As seen in other contributions before, A Decision-Theoretic Perspective on Machine Learning plays an important role also here, making this contribution a perfectly natural fit for the present habilitation thesis. \\

Let's first look at a little more background: PL is a method of \textit{semi-supervised learning (SSL)} (see, e.g., \cite{ChapelleSSL}). Roughly speaking, SSL is about improving the learning of (most commonly) a classifier by using, in addition to a (usually comparatively small) labeled data set, a (usually comparatively large) unlabeled data set. Here, PL uses the unlabeled data in a very specific way: it predicts labels of unlabeled data with a model trained on labeled data and adds certain predictions, then called \textit{pseudo-labels}, to the training data. This process is then repeated until some predefined stopping rule applies. Clearly, for making a decision about which of the pseudo-labels should be added to the training data in each step, a criterion for  \textit{pseudo-label selection (PLS)} is needed. While most traditional criteria for PLS rely on one single model (see, e.g., \citet{triguero2015self,rizve2020defense}), and even on the concrete fitted model, Contribution 7 intends to make the process of PLS more robust by considering multiple initial models.\footnote{Note that Contribution 7 covers more approaches for robustification of PLS, concretely \textit{multi-label} (Section 3.2) as well as \textit{multi-data} (Section 3.3) approaches, but these are left out for reasons of brevity here. Instead, we focus on the main line of research of the contribution and refer the reader to the corresponding sections for details on these approaches.}  \\

For this, we proceed as follows: The paper builds directly on the framework proposed in \citet{pmlr-v216-rodemann23a}, which formulates PLS as a \textit{decision problem} and proposes to select the pseudo-labels with respect to the \textit{Bayes-criterion} with respect to some predefined prior distribution. While this approach robustifies the PLS against the initial model \textit{fit} (by incorporating information from all potential parameter settings of the considered model), our article goes one step further and proposes robustifications against the initial model \textit{choice}. We achieve this as follows: Instead of using only one model to predict the pseudo-labels, we now use several. Each of these models leads to a separate decision problem for PLS in the sense of \citet{pmlr-v216-rodemann23a}, each with its own utility function (again in the sense of \citet{pmlr-v216-rodemann23a}). The question now is how to deal with the various utility functions. We propose various approaches here: In Definition 3 of Contribution 7, the decision problem of PLS is directly equipped with a multidimensional utility function, composed of those of the separate decision problems.\\

The first approach treats the induced problem as a decision problem with a \textit{multi-dimensional} utility function. As the single dimensions, i.e., the different utility functions arising from the different models considered, have a clear cardinal interpretation, the range of the multi-dimensional utility function can be embedded into a preference system in the spirit of the one defined in Equations (\ref{cps}) - (\ref{cps2}) of the present thesis. This makes the problem of PLS a decision problem under weakly structured information in the sense of Section~\ref{foundations} of this text. Accordingly, an appealing criterion for approaching this decision problem is via the GSD-relation as recalled in Definition~\ref{GSD}, where $\mathcal{M}=\{\Tilde{\pi}\}$ is chosen to consist only of the precise prior distribution $\Tilde{\pi}$ on the extended state space $\Tilde{\Theta}$ (see~\citet[Section 3.1.1]{pmlr-v215-rodemann23a}): In every step, we add exactly those pseudo-labels to the training data, which are not (strictly) GSD-dominated by any other available pseudo-label. This approach then also enables a completely natural generalization (see Section 4.3 of the contribution): If the prior distribution $\Tilde{\pi}$ cannot be specified precisely, e.g. if the independence of the models involved cannot be guaranteed, the GSD approach can be generalized directly to a credal set. For this, we just replace $\{\Tilde{\pi}\}$ by the corresponding credal set and proceed as before.\\

The second approach for treating the multi-dimensional utility of the PLS decision problem consists in transforming it into a weighted sum (see Definition 4 of the contribution). After a brief discussion about the fact that such a weighted sum of individual utilities should only be used if a natural method for determining the weights is available, we turn to a situation in which exactly this is the case: nested models (see Section 3.1.2 of the contribution). Here, for example, the weights can be used to penalize the complexity of the models involved, which can naturally be operationalized by the proportion of the parameters used to the number of parameters in the complete model. For this case, we propose a modified Bayes criterion (see Definition 5) and an algorithm for its evaluation (see Algorithm 1). \\

Next, in Section 4 of Contribution 7, we consider another way of robustifying PLS: As mentioned before, \citet{pmlr-v216-rodemann23a} proposes to solve PLS by using a Bayes-criterion in order to mitigate the influence of the original fitted model for determining the pseudo-labels on the process of PLS. While this approach is very convincing if a suitable prior is available, it may even prove counterproductive in the case of an unsuitable prior. If the prior knowledge about the distribution of the model parameters is therefore only unspecific, it is better to fall back on a generalized priori evaluation, in our case a credal set $\mathcal{M}$. In order to also incorporate the information encoded in the likelihood function, we then propose the following procedure: We first compute the subset $\Tilde{\mathcal{M}} \subseteq \mathcal{M}$ of compatible priors that produce a marginal likelihood which is still at least as high as $\alpha$-times the maximum possible marginal likelihood among all $\pi \in \mathcal{M}$. Here $\alpha \in [0,1]$ is a parameter controlling how strong the influence of the likelihood should be.\footnote{This update method for credal sets was proposed in \citet{cattaneo2013likelihood,cattaneo2014continuous} in a very general setting, and considered and translated to the context of objective Bayesianism under the term \textit{soft revision} in \citet{comment}.} Afterwards, we select pseudo-labels according to the $\Gamma$-maximin criterion (see, e.g., \citet{troffaes_ijar}) with respect to the updated credal set~$\Tilde{\mathcal{M}}$: For each pseudo-label, we compute the lowest possible expected utility compatible with the updated credal set and select (one of) the pseudo-label(s) receiving maximum worst-case expected utility. \\

Section 5 of Contribution 7 demonstrates the power of (some of) the proposed method in terms of both simulated and real-world data, where they show very promising results (compare Section 6 of the contribution). Note that the methods relying on credal sets are not included in the experiments as, at the time of the paper was published, methods for their evaluation were still lacking. We refer the reader to Section~\ref{frb} of this thesis for a brief update on this situation. 
\bigskip
\begin{mybox}
\begin{center}
    {\bfseries{Contribution 8}}
\end{center}
Christoph Jansen, Georg Schollmeyer, Julian Rodemann, Hannah Blocher, Thomas Augustin (2024): Statistical Multicriteria Benchmarking via the GSD-Front. Forthcoming in: \textit{Advances in Neural Information Processing Systems 37 (NeurIPS 2024).} The final version is available at: \url{https://openreview.net/forum?id=jXxvSkb9HD}\\

\begin{center}
    {\bfseries{Original Abstract}}\\
\end{center}

\textit{Given the vast number of classifiers that have been (and continue to be) proposed, reliable methods for comparing them are becoming increasingly important. The desire for reliability is broken down into three main aspects: (1) Comparisons should allow for different quality metrics simultaneously. (2) Comparisons should take into account the statistical uncertainty induced by the choice of benchmark suite. (3) The robustness of the comparisons under small deviations in the underlying assumptions should be verifiable. To address (1), we propose to compare classifiers using a generalized stochastic dominance ordering (GSD) and present the GSD-front as an information-efficient alternative to the classical Pareto-front. For (2), we propose a consistent statistical estimator for the GSD-front and construct a statistical test for whether a (potentially new) classifier lies in the GSD-front of a set of state-of-the-art classifiers. For (3), we relax our proposed test using techniques from robust statistics and imprecise probabilities. We illustrate our concepts on the benchmark suite PMLB and on the platform OpenML.}
\end{mybox}
\bigskip
Contribution 8 (see \citet{jansen2024statisticalmulticriteriabenchmarkinggsdfront}) follows-up on Contribution 5 of this habilitation project and further develops and substantially extends the GSD-based approach to classifier comparison. In particular, the ideas from Contribution 10 (see Part C of the present thesis) for robustifying statistical testing using ideas from imprecise probabilities as well as the possibility of including several, potentially differently scaled, quality metrics simultaneously play major roles here. The main extensions consist in extending our test to the multi-classifier setting as well as in providing detailed theoretical results in the context of statistical inference. Specifically, the substantial extensions compared to Contribution 5 are in the following respects (compare also Section 1.2 of Contribution 8):\\

\textit{GSD-Front:} We present the concept of the GSD-front, i.e., the set of not strictly dominated classifiers with respect to the GSD-relation, as an information-efficient improvement of the Pareto-front that nevertheless avoids too strong assumptions on scale or importance on the involved metrics. The improvement in utilizing the available information compared to a classical Pareto-analysis is that two additional sources of information can be included: First, by modeling via a preference system (as already seen in the summary of Contribution 5), the partial cardinal information contained in the cardinally interpretable performance metrics can also be included. The main idea is here that, if there is no potential conflict in the purely ordinal metrics, the cardinal ones can be used to model partial quality intensity by means of differences. Secondly, because GSD is an expected value-based criterion, information about the distribution of the data sets can also be included, where, in case only samples are available, the distributional information can still be approximated by the corresponding empirical analog. The result is a much more information-efficient framework than under Pareto analysis.\\

\textit{Estimation:} In Section 3, Definition 3.3 ii), we propose a set-valued estimator for the GSD-front, the $\varepsilon$-\textit{empirical GSD-front}, which allows us to approximate the true GSD-front based on \textit{i.i.d.} benchmark suites. We then provide sufficient conditions for the strong consistency of this estimator, which are based on considerations from statistical learning theory, more precisely Vapnik-Chervonenkis theory (compare Theorem 3.6). Importantly, Theorem 3.6 is non-trivial in the sense that the assumption of a finite Vapnik-Chervonenkis dimension can also fail to be satisfied. However, we show (see Corollary 5.1 of the Contribution 8) that this assumption is satisfied in the prototypical benchmarking problem considered in the paper. Furthermore, in Theorem 3.8 of the paper, we show that our estimator is both compatible with and generally more informative than the underlying Pareto-front. \\

\textit{Testing:} In Section 4 of Contribution 8 we turn to the question of statistical testing. More precisely, we develop (static and dynamic) statistical (permutation-)tests for checking if a classifier is in the GSD-front of some set $\mathcal{C}$ of state-of-the-art classifiers (corresponding to the static version) or in the GSD-front of some subset of this set $\mathcal{C}$ (corresponding to the dynamic version). In contrast, the test proposed in Contribution 5 was constructed to test another alternative hypothesis of actual interest, namely that a classifier $C$ dominates another classifier $C'$ w.r.t. GSD. Because a valid statistical level-$\alpha$-test with such a choice of hypotheses is not reachable (e.g., \citet[p. 206]{whang2019econometric}), Contribution 5 had to leave the proper Neyman-Pearson testing framework, exchanging the roles of the null hypothesis and the alternative hypothesis. Now, in contrast, the approach in Contribution 8 is directly interested in the GSD-front (and the above considerations were one of the reasons why we developed the new conceptualization of a test that uses the GSD-front). Here, the logic of testing changes in a way such that our test(s) is/are not only valid level-$\alpha$-tests, but, (compared to Contribution 5, where the test is only sensitive to directed alternatives) they are also \textit{consistent} tests for the alternative of a classifier of being in the GSD-front.\\

\textit{Different conceptualizations of multidimensional performance metrics:} Another important original contribution of our paper is of a conceptual rather than technical nature: the distinction between two fundamentally different motivations to use multidimensional quality metrics (compare Section 1.1): \textit{Performance as a latent construct} on the one hand versus\textit{ Quality as a multidimensional concept} on the other hand. While the former interprets multidimensional quality metrics as a set of metrics operationalizing a \textit{latent} quality construct, the latter chooses different metrics to trade-off between several quality dimensions that appear relevant to the problem at hand. This conceptual distinction also finds its way into our applications. In particular, for the OpenML dataset we analyze different quality metrics (accuracy and runtime) that are clearly measuring different inherent dimensions of performance. Furthermore, for the PMLB benchmark suite, we developed a notion of \textit{robust accuracy} as a multidimensionally operationalized latent concept that is of its own interest.\footnote{Note that our approach differs from that of \citet{zhu:noise:2004} in the sense that we do not sample from the support of the distributions but instead directly from the empirical distributions. This may be seen as an advantage because this guarantees that all marginal distributions of the perturbated variables are the same as the original, unperturbated distributions.}\\

\textit{Robustification:} Based on the ideas for robust testing developed in Contribution 10 of the present habilitation thesis, we propose a method for quantifying how robust the test decisions of our proposed test are under deviations from the underlying assumption of \textit{(i.i.d.)}-samples. The idea is based on the theory of imprecise probabilities: In Section 4.2 of the contribution, we deliberately perturb the empirical measures the permutation tests are based on by using contamination models. We then investigate, up to which contamination degree the found classifier ranking remains to be significant. Again, we propose our robustified testing framework in a static and a dynamic variant.\\

\textit{Applications and Implementation:} Opposed to Contribution 5, we consider experiments with \textit{mixed-scaled} (ordinal and cardinal) multidimensional quality metrics, instead of only cardinal ones. This allows to incorporate also truely ordinal criteria like, e.g., runtime levels or robustness w.r.t. class and attribute noise. Finally, again opposed to Contribution 5, the experiments are run over two benchmark suites (PMLB and OpenML), rather than just a selection of 16 data sets. This is possible thanks to a more efficient implementation that is freely available and easily adaptable to comparable problems.\\

From my point of view, when perceived on a more general level, the concepts (further) developed in Contribution 8 have the potential to provide the basis for a completely new approach to \textit{reliable} machine learning. This is for the following reasons: 1.) The possibility to model data with non-standard scales of measurement is much more align with real-world machine learning problems than most of the over-idealized models commonly used. 2.) Introducing inferential concepts allows for obtaining general error bound rather than only empirical results. 3.) The possibility of analyzing test decisions under deviations from the \textit{i.i.d.}-assumption (which is often not met/not checkable in practice) allows for judging the robustness of results.
\subsection{Future Research on Part B} \label{frb}
As I did at the end of Part A, I will now outline some perspectives for future research in connection with the contributions to Part B. Again, I make no claim to completeness: the points of reference mentioned are simply those that seem most relevant at the time of writing. Perhaps the most direct possibility for a follow-up is offered by Contribution 7: As described in the corresponding section, various possibilities were proposed here to robustify the method of pseudo-label selection (PLS) from semi-supervised learning (SSL). While most of them were also investigated experimentally, the discussion of methods based on credal sets remained on a purely theoretical level. This was not least due to the fact that the Laplace approximation used to evaluate the integrals associated with the Bayesian criterion (see~\citet{pmlr-v216-rodemann23a}) was not (directly) transferable to the proposed $\Gamma$-maximin criterion under soft revision. \\

However, in a recent bachelor thesis supervised by our group, a possibility was found to also approximate the infimal integral associated with the $\Gamma$-maximin criterion using a nested Laplace approximation.\footnote{The thesis was written by Stefan Dietrich, under the supervision of Julian Rodemann. The idea for the nested approximation is due to Stefan Dietrich. A substantially extended version of (parts of) the thesis are the basis of a paper, which is accepted for the proceedings of SMPS 2024. The order of authors for this paper is Stefan Dietrich, Julian Rodemann, and Christoph Jansen.} With the help of this approximation technique, the methods for PLS based on credal sets can now also be investigated experimentally. Initial results in this direction look very promising: The $\Gamma$-minimax criterion under soft revision based on the multi-model utility from \citet[Definition 4]{pmlr-v215-rodemann23a} outperforms all state-of-the-art methods in many relevant settings. We hope to be able to present detailed experimental studies soon, which will confirm the positive initial impressions described above.\\

Further exciting research perspectives arise in connection with Contributions 5 and 8: First of all, it should be recalled that Contribution 8 is already an extension of Contribution 5. As discussed in more detail before, this extends the GSD-test introduced in Contribution 5 from the two classifier cases to a test in connection with the GSD-front (compare the summary of Contribution 8 for details). A natural next step would be to transfer the comparison framework presented here to situations other than the comparison of classifiers. As already noted in the outlook of Contribution 8, none of our concepts depends on the objects to be compared being classifiers. On the contrary, any situation in which objects are to be compared based on different (potentially differently scaled) metrics over a random selection of instances can be analyzed using these ideas. For instance, applications of our framework to the multicriteria deep learning benchmark suite \texttt{DAWNBench} \citep{coleman2017dawnbench} or the bi-criteria
optimization benchmark suite \texttt{DeepOBS} \citep{schneider2018deepobs} appear straighforward. Interestingly, note that a similar transfer of the depth-based framework for classifier comparison proposed in Contribution 6 to the latter paper was recently carried out in \citet{rodemann2024partial}.\\

Another, rather methodical, research perspective in the context of Contributions 5 and 8 are extensions of the framework to regression-type analysis: In its current form, analyses based on the GSD-front do not account for meta properties of the data sets in the respective benchmark suite. A straightforward extension to the case of additional covariates for the data sets is to stratify by these for the GSD-comparison. This would allow for a \textit{situation-specific} GSD-analysis, presumably yielding more informative results. As a simple example, for each data set, one could include a covariate storing the scientific field the data set is originating from. Then, a GSD-based analysis could be used to construct and test \textit{field-specific} classifier rankings, which would presumably allow for more structured statements than comparisons across several different fields. Many other different meta properties could be straightforwardly included as well.\\

Finally, there are of course also numerous promising follow-up opportunities in connection with Contribution 6: One is the transfer of the framework to other scenarios, as mentioned above, for example in \citet{rodemann2024partial,rodemann2024reciprocallearning}. Another is the inferential further development of Contributions 6 and 9, as described in more detail in the outlook for Part C, see \ref{frc}. 
\newpage
\section{Robust Statistics under Non-Standard Scales of Measurement}
\noindent 
\\
\begin{minipage}{0.3\textwidth}

\end{minipage}
\hfill
\begin{minipage}{0.5\textwidth}
\textit{In my opinion the crucial attribute of robust methods is stability under small perturbations of the model. I am tempted to claim that robustness is not a collection of procedures, but rather a state of mind [...].}\\[.2cm] \hfill -----------------------------------------(\citet[p.1250]{Huber2011})
\end{minipage}\\[.5cm]
I now come to the third and last main part of this habilitation thesis: the application of decision theory under weakly structured information to robust statistics under non-standard scales of measurement. I will start by carefully demonstrating how the decision-theoretic concepts recalled in Part A of this thesis can be transferred to this setting. Afterwards, I will give brief general summaries of the two contributions building the core of Part C. Finally, I elaborate on promising avenues for future research projects based on the work presented.
\subsection{Non-Standard Scales of Measurement}
The following general introduction will be a little shorter than those for Parts A and B. The main reason for this is that a major part of the work has already been done in the introduction to Part B: The transfer of the originally purely decision-theoretically motivated concept of a preference system (see, e.g., Definition \ref{ps} in the present thesis) to scenarios of machine learning and statistics. In the following, I will therefore only deal with the specifics of a reinterpretation of preference systems in connection with \textit{robust statistical analysis}. I will also briefly explain how random variables whose values correspond to partial orders fit into this picture.
\subsubsection{Scale-Robust Stochastic Orders}
Let us first recall what the main motivation here was: preference systems can be used to model data (or more generally sample spaces) with \textit{locally varying scale of measurement}. The relations $R_1$ and $R_2$ in the definition of a preference system $\mathcal{A}=[A, R_1,R_2]$ play a key role in this transfer: As already explained in Part B, $R_1$ is used to model the available ordinal, and $R_2$ is used to express the available cardinal information. Intuitively, we can then say that
\begin{itemize}
    \item $A$ is \textit{locally almost cardinal} on subsets where $R_1$ and $R_2$ are very dense. Any analysis of the variable $X$ should be based on the set $\mathcal{U}_{\mathcal{A}}$ alone, which reflects this local property by implicitly allowing the compatible candidate scales low variability on the dense parts. 
    \item $A$ is \textit{locally at most ordinal} on subsets where $R_2$ is sparse or even trivial. Any analysis of the variable $X$ should be based on the set $\mathcal{U}_{\mathcal{A}}$ alone, which reflects this local property by implicitly allowing the compatible candidate scales high variability on the sparse parts.
\end{itemize}
In the context of statistics and in the light of the reinterpretation of preference systems just recalled, each representation $u \in \mathcal{U}_{\mathcal{A}}$ can then be interpreted a \textit{cardinal candidate scale} of the sampling space $A$ under consideration. Of particular interest for this work (and also for statistics in general) are \textit{stochastic orders} between random variables with values in $A$, which are robust under the choice of the concrete candidate scale from $ \mathcal{U}_{\mathcal{A}}$. If a unique cardinal scale $u^*:A \to \mathbb{R}$ is available\footnote{That is, if the class $\mathcal{U}_{\mathcal{A}}$ consists exactly of the positive linear transformations of $u^*$.} and we know the variables of interest $X,Y: \Omega \to A$ map from some space $\Omega$ for which a probability measure $\pi$ is available,\footnote{Note that, again, we leave measurability issues aside, compare Footnote~\ref{nontechnical}. The detailed and technical rigorous definitions can be found in the respective contributions.} then the most popular stochastic ordering, the \textit{expectation ordering} $\succsim_{E(u^*)}$ with respect to the scale $u^*$,
is obtained by setting $(X,Y) \in \succsim_{E(u^*)}$ if and only if
\begin{equation} \label{EU}
    \mathbb{E}_{\pi}(u^* \circ X)=\int_{\Omega}u^* \circ X d\pi\geq \int_{\Omega}u^* \circ Y d\pi= \mathbb{E}_{\pi}(u^* \circ Y).
\end{equation}
Here, random variables are ranked according to the expectations of their numerical \textit{certainty equivalents} induced by the scale $u^*$. However, this expectation order is generally no longer well-defined if $\mathcal{U}_{\mathcal{A}}$ cannot be represented as a class of positive linear transformations of a fixed candidate scale: In general, it might very well be the case that $(X,Y) \in \succsim_{E(u_1)}$ and at the same time $(Y,X) \in \succ_{E(u_2)}$, where both $u_1,u_2 \in \mathcal{U}_{\mathcal{A}}$ represent the same underlying preference system. This makes it clear that the expectation order between random variables depends very strongly on the correct choice of the cardinal scale and thus even weak perturbations of this scale can turn the comparison of interest into the complete opposite.\\

As a consequence of this consideration, stochastic orders are desirable which are robust to the specification of the cardinal scale. The easiest way to robustify the expectation order, which is basically a convincing idea, is to simply require it for all compatible candidate scales: We can define $(X,Y) \in \succsim_{E(\mathcal{U}_{\mathcal{A}})}$ if and only if
\begin{equation} \label{GEU}
 \forall u \in  \mathcal{U}_{\mathcal{A}}:~~  \mathbb{E}_{\pi}(u \circ X)=\int_{\Omega}u \circ X d\pi\geq \int_{\Omega}u \circ Y d\pi= \mathbb{E}_{\pi}(u \circ Y).
\end{equation}
The expectation order generalized in this way is then, by construction, independent of, and thus robust to, the choice of the specific cardinal scale. Of course, this robustness comes at a price: the generalized expectation order is only a preorder, i.e. it generally has incomparable elements. This is exactly where the special importance of preference systems comes into play in this context: The relation $R_2$ also makes it possible to incorporate local/partial metric/cardinal knowledge. This guarantees that the set can be counted as small as possible (and still be information-valid). This also minimizes the loss of ordering power associated with generalization.\\

Finally, a mathematically obvious but substantively all the more important note: The generalized expectation order $\succsim_{E(\mathcal{U}_{\mathcal{A}})}$ from Equation (\ref{GEU}) corresponds exactly to the GSD-relation $R_{(\mathcal{A},\{\pi\})}$ from Definition~\ref{GSD} in the present habilitation thesis, i.e., the GSD-relation based on a classical probability $\pi$ rather than a non-singleton credal set. The considerations just repeated have thus provided an alternative motivation, completely detached from decision theory, for considering this very relationship: The scale-robust comparison of random variables that map into non-standard sample spaces.
\subsubsection{Robust Statistical Testing} \label{robtest}
Another important question for the present work is the search for suitable statistical tests for the scale-robust stochastic orders just discussed. In other words: Given only \textit{i.i.d.} samples $\mathbf{X}=(X_1, \dots , X_n)$ and $\mathbf{Y}=(Y_1, \dots , Y_m)$ of the variables $X$ and $Y$ of interest, when can we, with a certain error probability, conclude from this information that $X$ and $Y$ ordered with respect to the generalized expectation order (or, equivalently, the GSD-relation) $R_{(\mathcal{A},\pi)}$? An answer to this question is given in Contribution 10 of this cumulative habilitation thesis (and outlined in the following summary of the article), where we use non-parametric permutation-based statistical tests to tackle this problem.\\

\begin{figure}
    \centering
    \includegraphics[angle=90,width=.46\linewidth]{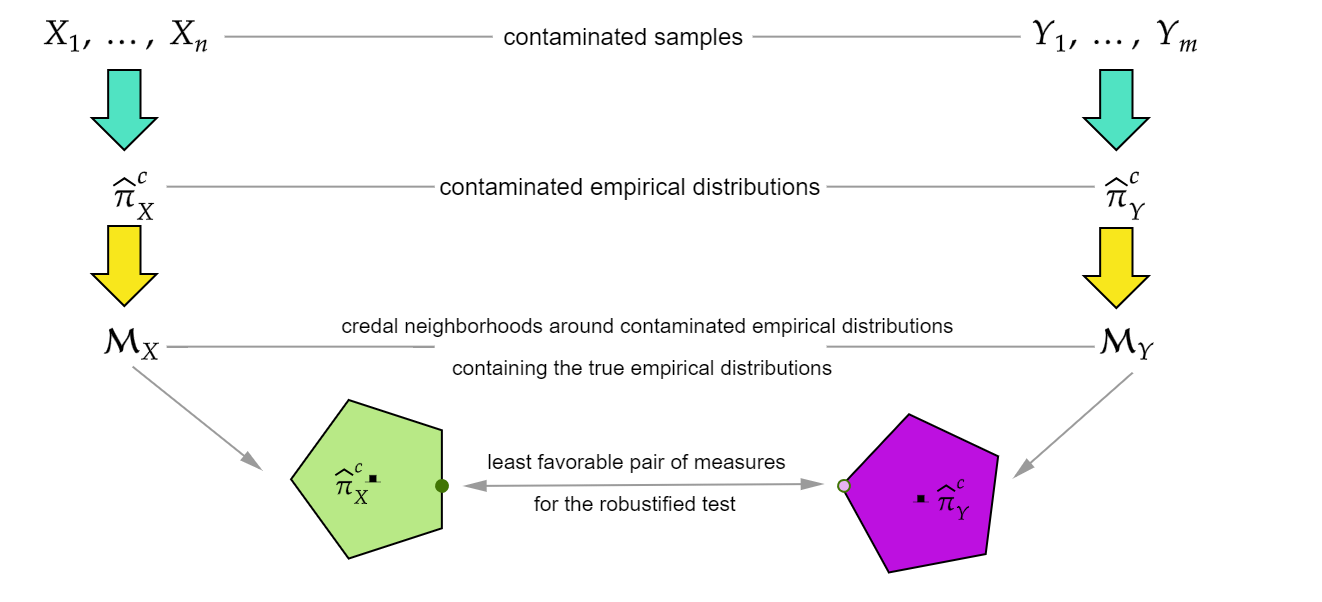}
    \caption{The diagram shows a schematic visualization of how the robustification of the (permutation-based) statistical testing under deviaions from the i.i.d. assumption, e.g., used in Contributions 8 and 10, works. Roughly spoken: If we start with samples of $X$ and $Y$ which are contaminated in the sense that they were not necessarily sampled i.i.d. (e.g., think that $k_1 <n$ resp. $k_2 <m$ of the samples of $X$ resp. $Y$ origin from some arbitrary but unknown distribution), this will lead us to contaminated empirical distributions of both variables. If we can, however, specify credal sets around those contaminated measures for which we know the true empirical measures have to be contained in them (e.g., think that we can say that a maximum of $\tfrac{k_1}{n}$ resp. $\tfrac{k_2}{m}$ of the probability mass is arbitrarily distributed), then we can perform the (in our case permutation-based) statistical test for the \textit{least favorable pair} of measures from these two credal sets, i.e. the combination of candidate emprirical measures that make it most difficult to reject the null hypothesis. This procedure yields a valid (yet conservative) test for the predefined significance level.}
    \label{scheme_test}
\end{figure}

The same article also looks at another type of robustness. While we have so far focused on robustness to the choice of scale, another form comes into play in the context of statistical tests: robustness to the \textit{i.i.d.-assumption} of the samples $\mathbf{X}=(X_1, \dots , X_n)$ and $\mathbf{Y}=(Y_1, \dots , Y_m)$ on which the validity of our proposed statistical  (permutation-based) test depends. Specifically, the following question arises here: How strongly does the significance of a test decision depend on the independent and identical distribution of the variable samples?  This question is also addressed in detail in Contribution 10, where the theory of imprecise probabilities (and of credal sets in particular) plays a decisive role to model the amount of contamination of the i.i.d. case. There is also an interesting cross-connection between the different parts of this work in this context: The considerations on robust testing made in Contribution 10 are extended to the multi-variable case in Contribution 8 of Part B (in a different context). A schematic visualization of the robustified test procedure under deviations from the i.i.d. assumption is given in Figure~\ref{scheme_test}.
\subsubsection{Partial Orders as Data}
So far, in our introductory remarks to Part C, we have mainly considered one type of non-standard data, namely variables that map into preference systems. However, another type of non-standard data plays an important role for the following considerations in this part of the thesis (especially in Contribution 9): random variables whose \textit{values are partial orders}. Here it is useful to remember that we have already looked at the same type of data in Contribution 6. In purely formal terms, of course, this type of variable is also one that maps into preference systems, i.e. a special case of the first type of non-standard data: The set of all partial orders is itself partially ordered by set inclusion. If we define the set inclusion as $R_1$ and choose $R_2$ as trivial preorder, we again obtain variables with values in preference systems. However, such an analysis, although formally completely correct, would be of little use here, since without cardinal information one would quickly be reduced to a consideration of first-order stochastic dominance. As already seen in Contribution 6, an analysis based on the theory of data depth proves to be more productive here.
\subsection{Contributions to Part C}
I will now briefly summarize the two publications forming the core of Part C of the present work. The order in which the publications are discussed is chronological by date of publication, starting with:
\bigskip
\begin{mybox}
\begin{center}
    {\bfseries{Contribution 9}}
\end{center}
Hannah Blocher, Georg Schollmeyer, and Christoph Jansen (2022): Statistical models for partial orders based on data depth and formal concept analysis. In: Ciucci, D.; Couso, I.; Medina, J.; Slezak, D.; Petturiti, D.; Bouchon-Meunier, B.; Yager, R.R. (eds): \textit{Information Processing and Management of Uncertainty in Knowledge-Based Systems.} Communications in Computer and Information Science, 1602: 17-30, Springer.\\

\begin{center}
    {\bfseries{Original Abstract}}\\
\end{center}

\textit{In this paper, we develop statistical models for partial orders where the partially ordered character cannot be interpreted as stemming from the non-observation of data. After discussing some shortcomings of distance based models in this context, we introduce statistical models for partial orders based on the notion of data depth. Here we use the rich vocabulary of formal concept analysis to utilize the notion of data depth for the case of partial orders data. After giving a concise definition of unimodal distributions and unimodal statistical models of partial orders, we present an algorithm for efficiently sampling from unimodal models as well as from arbitrary models based on data depth.}
\end{mybox}
\bigskip
Similar to Contribution 6 (see Part B), Contribution 9 (see \citet{bsj2022}) is dedicated to random variables whose range is given by the set of all partial orders, i.e., data for which every single observation is a partial order. Completely natural applications in which this type of non-standard data occurs are, for example, the comparison of algorithms with regard to several quality measures simultaneously, in which a (potentially) different partial order arises between the algorithms for each data set under consideration (compare Contributions 5, 6 and 8), or the comparison of different universities using survey data, which also explicitly contain an option for ``non-comparability" (see, e.g., \citet{katzen,Schollmeyer_isipta}). In sharp contrast to Contribution 6, which is concerned with the \textit{descriptive} analysis of samples of such random variables, Contribution 9 focuses on \textit{statistical models} of the distribution of these random variables.\\

Of course, Contribution 9 is not the first paper to deal with stochastic models over the set of all possible (partial) orders. On the contrary, research into models for \textit{total} orderings, or \textit{rankings}, has a long standing tradition, where particularly prominent examples that have intensively been followed-up on are given by the Thurstone model~\citep{thurstone}, the Bradley-Terry model~\citep{bradley1952rank}, or the Mallows model~\citep{mallows1957non}. Stochastic models for \textit{partial} orders, which are also addressed in Contribution 9, have also attracted less, but still considerable, interest: For instance, \citet{davidson70} generalizes the Bradley-Terry model to partial orders, \citet{brandenburg} define models based on a generalized version of Kendall's $\tau$, and \citet{chierichetti2018mallows} adapt the Mallows model to account for top-k orders. For a textbook on the topic, see \citet{Critchlow1985}.\\

However, our work differs significantly from the latter approaches. While most existing approaches to stochastic modeling of partial-order-valued random variables (implicitly) assume\footnote{The main point here is that most of these models are based on generalized distances between total orders, compare the discussion in Section 2 of Contribution 9.} that the involved partial orders represent deficiently observed total orders, we explicitly conceive them as entities on their own: If the objects $a$ and $b$ are incomparable in one of the observed partial orders, we do not interpret this as a lack of information, but rather as information that is just as valid as the comparability of objects. Note that there is a parallel here to the distinction between the ontic and epistemic view, as is made in the area of set-valued random variables, so-called random sets (see, e.g., \citet{Couso_2014}): While our interpretation of partial order is more consistent with the ontic view (``precise observations of something genuinely imprecise''), many of the existing models are more consistent with an epistemic view (``imprecise observations of something genuinely precise'').\\

Technically, we achieve this by -- roughly speaking -- not building models based on generalized distances between total orders, but by completely replacing the involved distances by suitable \textit{depth functions} (see the discussion of Contribution 6 in Part B of this thesis) on the set of all partial orders. In a nutshell, depth functions, originally introduced for $\mathbb{R}^n$-valued data, measure the centrality or outlyingness of a data point with respect to a data cloud (\textit{empirical depth functions}) or a probability distribution (\textit{theoretical/population-based depth functions}) \citep{tukey75,liu90,SERFLING2004,serfling2006depth}. In order to be able to formalize depth functions in this abstract framework in a meaningful way and to formulate desireable properties of the distributions based on these depth functions, in Section 3, we use techniques from \textit{formal concept analysis} (e.g., \citet{ganter12}). Roughly speaking, formal concept analysis powerfully offers the possibility of a \textit{purely relational} (as opposed to numerical) data analysis, which is based on an abstraction of the concept of the cross-table, as is absolutely common in descriptive statistics, and enables very general statements to be made even about the analysis of non-standard data. Specifically, in Section 4 of the paper, we define a suitable \textit{formal context} (compare the paper's Section 3) over the set of all partial orders, which equally takes into account both the comparability and incomparability of objects, therefore guaranteeing an ontic view on the inolved partial orders as discussed above. \\

In the paper, we then consider three different depth functions for this formal context: the \textit{generalized Tukey's depth} as proposed in \citet{schollmeyer17a} as a generalization of \citet{tukey75}, the \textit{peeling depth} and \textit{enclosing depth} (for both, see Definition 2) proposed in our paper. Intuitively speaking, the peeling depth successively removes the extreme points\footnote{For the exact definition of an extreme point in this abstract context, I refer the reader to Footnote 2 of Contribution 9.} (i.e., the most outlying points) of the data cloud from the outside inwards and then, increasing in the number of iterations, assigns the same depth value to all points within the same exclusion level. Contrarily, the enclosing depth is based on the exact opposite intuition: Here, starting from a fixed central order (which basically serves as a model parameter), subsets of partial orders are successively added to the data cloud from the inside to the outside, whereby the orders included in the same stage are each assigned the same depth value falling in the number of iterations.  \\

Based on these three depth functions, we then discuss depth-based stochastic models over the set of all partial orders that ``parametrically'' depend only on a fixed central, i.e. deepest, partial order (see Equation (1) in the paper), therefore, generalizing the notion of location families to depth-based stochastic models. In particular, we are interested in the conditions under which non-trivial \textit{unimodal} distributions on the set of all partial orders can be generated on the basis of these depth functions, and how strongly the models induced by the different depth functions differ in their discriminatory power, i.e., how many different depth values can be actually attained by them. Finally, in Section 6 of the paper, we present an algorithm that can be used to simulate partial orders from depth-based stochastic models of the form as given in Equation (1) and discuss the complexity class. The presented algorithm is based on the \textit{acceptance-rejection method} discussed in \citet{ganter2011random}. The possibility of sampling from the proposed models then enables simulation studies and inferential analyses to be carried out within this abstract framework. Some very preliminary thoughts on this aspect can be found in the research outlook for Part C of this thesis.
\bigskip
\begin{mybox}
\begin{center}
    {\bfseries{Contribution 10}}
\end{center}
Christoph Jansen, Georg Schollmeyer, Hannah Blocher, Julian Rodemann, and Thomas Augustin (2023): Robust statistical comparison of random variables with locally varying scale of measurement. In: Evans, R.; Shpitser, I. (eds): \textit{Proceedings of the Thirty-Ninth Conference on Uncertainty in Artificial Intelligence.} Proceedings of Machine Learning Research, 216: 941-952, PMLR.\\

\begin{center}
    {\bfseries{Original Abstract}}\\
\end{center}

\textit{Spaces with locally varying scale of measurement, like multidimensional structures with differently scaled dimensions, are pretty common in statistics and machine learning. Nevertheless, it is still understood as an open question how to exploit the entire information encoded in them properly. We address this problem by considering an order based on (sets of) expectations of random variables mapping into such non-standard spaces. This order contains stochastic dominance and expectation order as extreme cases when no, or respectively perfect, cardinal structure is given. We derive a (regularized) statistical test for our proposed generalized stochastic dominance (GSD) order, operationalize it by linear optimization, and robustify it by imprecise probability models. Our findings are illustrated with data from multidimensional poverty measurement, finance, and medicine.}
\end{mybox}
\bigskip
Contribution 10 provides a detailed investigation of the GSD-order $\succsim_{E(\mathcal{U}_{\mathcal{A}})}$ (or equivalently $R_{(\mathcal{A},\{\pi\})}$) as introduced in Equation (\ref{GEU}) (or Definition~\ref{GSD}) of the present thesis. In other words, the article is dedicated to the question of how statistical variable comparisons can be carried out under scale-robust stochastic orders. The main focus here is on three areas: (I) How can the GSD-order be tested statistically if only \textit{i.i.d.} samples of the variables to be compared are available? (II) How can the GSD order be characterized mathematically more precisely if more structural assumptions are made about the underlying preference system? (III) What are sensible real-world applications of the initially rather theoretical considerations of parts (I) and (II)?\\

The article first addresses question (I): Since we are in a completely non-parametric setting, the question of a suitable statistical test for checking whether two variables are ordered with respect to the GSD-relation is not easy to answer at first glance. Although meaningful test hypotheses can be established (see the introduction to Section 5 of the article) and a convincing test statistic can be specified (see Section 5.1), the distribution of this test statistic under the corresponding null hypothesis cannot be analyzed directly. For this reason, a permutation test is used (see, e.g., \citet{pg2012}). For this purpose, the fact that the test statistic used can be calculated using suitably selected linear programs is exploited (see Propositions 3 and 4 of Contribution 10). However, the selected permutation test, although statistically valid by construction, is potentially conservative for the test hypotheses under consideration: The null hypothesis of the permutation test, i.e. the equality in distribution of the variables being compared, is only the worst-case hypothesis of the null hypothesis that is actually of interest. To counteract this effect, we now propose to use a regularized version of the original test statistic. The idea of regularization here is based on the idea of parameter-driven regularization, as I repeated in Section \ref{reginpref} of this thesis.\\

Then, in Section 6 of Contribution 10, we address in detail the question of the robustness of the proposed test in the sense of Section~\ref{robtest} of this thesis. The concrete idea behind our robustification is that we allow the samples of the competing variables to be (potentially) biased. We assume that these biased samples are similar to the true ones in the sense that the associated true empirical laws are contained in the credal sets $\mathcal{M}_X $ and $\mathcal{M}_Y $ around the biased empirical laws, respectively. We then show how our proposed test can be modified to remain statistically valid under this contaminated scenario. Specifically, we investigate the situation in the case that both credal sets are contamination models (e.g., \citet{Walley.1991,mmd20120}) around the skewed empirical distribution of the samples. In this case, the test can be adapted by testing with a \textit{least favorable pair} of distributions from the corresponding credal sets instead of the empirical distributions (compare Propositions 5 and 6 of the contribution).\\

In Section 7 of Contribution 10, we then address the question (II) posed above: the characterization of the GSD-order for special preference systems. In doing so, we turn to a previously discussed class of preference systems: Multidimensional spaces with differently scaled dimensions, as recalled in Part B's Equations (\ref{cps}) - (\ref{cps2}). The main result here is Proposition 7 in Contribution 10, which (among other statements) establishes a characterization of GSD over the individual component orders if they are pairwise stochastically independent. Furthermore, Proposition 8 shows a precise characterization of the set of compatible representations of this special type of preference system if the underlying space has only one cardinal dimension. \\

Section 8 of the article then finally addresses the question (III) that we posed at the beginning. Specifically, we look at three different applications here. First, we turn to multidimensional poverty measurement (already mentioned in the Section \ref{mdsdsd} of this thesis). In accordance with the General Social Survey (ALLBUS) \citep{allbus-2014} we here account for three dimensions of poverty: income (numeric), health (ordinal, 6 levels) and education (ordinal, 8 levels), see also \citet{breyer2015skala}. The GSD-based comparison of the subgroups of men and women can be found in Section 8 of the article. As further applications we analyze a dermatology data set that contains variables on symptoms of the eryhemato-squamous disease, see \cite{demiroz1998learning} accessed via \cite{graf}, as well as the German credit data set that consists of variables on credit applicants, see \cite{graf}.
\subsection{Future Research on Part C}\label{frc}
As already declared for Parts A and B, also here what follows should not be confused with an exhaustive list of research perspectives concerning Part C. Instead, I roughly sketch three specific avenues for future research that seem most promising for me at the time I write the present section. The first directly concerns \textit{Contribution 9}. Recall that, in it, we proposed a \textit{statistical model} for random variables that take values in the set of all partial orders. Opposed to existing approaches (e.g., \citet{Critchlow1985,brandenburg}), this model is based on the concept of data depth rather than distances, therefore, as detailed in the contribution, relying on an \textit{ontic} view of the inserted partial orders rather than an \textit{epsitemic} one (see, e.g., ~\citet{Couso_2014} for a detailed clarification of these terms). \\

On the other hand, recall that Contribution 6 of the present thesis is also concerned with analyzing data consisting of partial orders, however, \textit{descirptively}. Taken together, Contributions 6 and 9 allow for i) descriptive analysis and ii) statistical modelling of this special type of non-standard data. In order to exploit the full spectrum of statistical analysis possibilities, the next natural step would be iii) \textit{statistical inference} for poset-valued random variables. Initial results can already be found in the articles mentioned above: While Contribution 9 provides an algorithm to sample from the proposed statistical models (compare Algorithm 1 in the contribution), Contribution 6 demonstrates that the empirical ufg depth converges almost surely against the associated population variant (compare Theorem 10 of the contribution). The next natural steps towards statistical inference would therefore be \textit{estimation} of model parameters from poset data, one- and two-sample \textit{tests} for poset data and \textit{regression} analysis for poset data.\\

The second promising avenue for future research in Part C again is based on the considerations in Contribution~9: Recall that one important aspect in the contribution is to treat the observed partial orders as entities on their own (i.e., give them an ontic interpretation) rather than corresponding to a deficient observation of an underlying total order (i.e., give them an epistemic interpretation). An interesting extension would now be to add additional epistemic uncertainty to the process: While still assuming that the objects of interest are \textit{partial orders} rather than total ones, we now allow certain edges of the observed partial orders to be not accessible. This adds additional epistemic uncertainty as for each such deficiently observed partial order a whole set of partial order extensions exists whose elements are equally plausible candidates for being the true underlying partial order. First steps for a meaningful treatment of such ontic data under epistemic uncertainty, or \textit{epiontic data} for short, were undertaken in \citet{Schollmeyer_isipta}. These first steps relied on the idea of a generalized form of \textit{cautious data completion} (e.g., \citet[Chapter 7]{augustin2014introduction}): Instead of computing the empirical depth just for one precisely observed sample of partial orders, we now compute it for every \textit{instantiation} of parial orders that is compatible with the corresponding sets of partial order extensions obtained by including epistemic uncertainty. Of course, as already computing the depth for a precisely observed sample of partial orders is computationally demanding (compare the contribution for a discussion of computational complexity), the same holds even more severe for the cautious data completion. However, as often only the lower and upper bound for the depth are of interest, the problem seems to be feasible at least in certain interesting situations. Concretely, in \citet{Schollmeyer_isipta}, we demonstrated that computing lower and upper depth values is doable for the generalized Tukey depth \citep{schollmeyer17a}. A natural next step is developing computational strategies for enabling similar analyses in the context of the ufg depth proposed in Contribution 6.\\

With regard to Part C, the last possible future research direction I want to elaborate on concerns extensions of Contribution 10. Here, mainly two aspects seem especially interesting:  In the contribution, we have layed a special focus -- for reasons of computational complexity -- to linear-vacuous models for robustifying the proposed permutation test. However, the idea of a priori identifying least favorable pairs of extreme points seems to generalize to any credals sets induced by belief functions in the sense of~\cite{s1976}. This would increase the chances of our methodology to be followed-up on, since the theory of belief functions is intensively used in practice in a broad range of scientific disciplines. The second aspect concerns the improvement of computational complexity: The LPs for checking in-sample GSD that are proposed in Contribution 10 become computer intensive for larger amounts of data. Although complexity already reduces drastically for the special case of preference systems arising from multi-dimensional structures with differently scaled dimensions (compare Section 7 of Contribution 10), Proposition 8 of the paper suggests that a further drastic reduction can be expected for only one cardinal dimension: The proposition gives a neat characterization of the set of compatible utility representations for this case, potentially allowing for a more efficient formulation of the respective linear programs in the contribution.
\newpage
\section*{Concluding Remarks}
\addcontentsline{toc}{section}{Concluding Remarks}
When I started working on this habilitation project, I had a concretely formulated goal in mind that could hardly have been more vague in its concrete implementation: Building on the considerations that formed the basis of my dissertation, I set out to develop a decision-theoretical foundation for certain problems of robust statistics and machine learning. But I wanted even more. I didn't want to be satisfied, as is often the case in literature, with reformulating well-known problems in these disciplines from a decision-theoretical perspective. A mere reformulation does nothing to solve a problem. I wanted to ensure that the transfer of decision-theoretical concepts to the challenges of the corresponding discipline would bring real added value, be it in the form of axiomatic justification of certain procedures or the solution of problems with the help of decision-theoretical criteria or algorithms.\\

My first starting point was very direct: I had the idea of using the preference systems we proposed in \citet{jsa2018} to improve query efficiency in expert and recommender systems. The intuition behind the idea was that the possibility of querying partial cardinal knowledge, i.e. partial information about preference intensity, and the existence of decision criteria that can also adequately process this type of information should pose major challenges for the corresponding state-of-the-art methods. And indeed, we were able to prove in the publication that the number of queries required to reach the optimal decision can be drastically reduced through the cooperation of partially cardinal information. In particular, we were able to show that the information encoded in the relation $R_2$ of the underlying preference system can be used to predict the most efficient query sequences. Given the increasing importance of the use of preferences in machine learning, the use of preference systems, and thus the utilization of partial cardinal information, could also prove to be extremely interesting in other areas. Concrete examples of this are \textit{inverse reinforcement learning}  \citep{ng}, where preference systems could offer a good trade-off between the extremes of perfect cardinal rewards and purely ordinal human feedback, \textit{deep learning} \citep{christiano2023deep}, where preference systems could be used to improve learning by allowing for more structured human feedback and \textit{language models} \citep{rafailov2023direct}, where preference systems could be used to stabilize the learning algorithms.  \\

My second starting point was somewhat more indirect. I posed myself the question: how can preference systems be used to improve statistical methods and make them more resistant to deviations from the assumptions on which they are based? We were able to provide an initial answer to the statistical part of the question in Contribution 10. Here we have transferred preference systems, which are originally motivated purely by decision-theoretical considerations, to statistical situations in which neither a clear ordinal nor a clear cardinal scale of measurement is given, but a hybrid of the two extremes, a \textit{locally varying} scale of measurement. The relation $R_1$ of the preference system was simply interpreted here as the available ordinal information in the data, the relation $R2$ was interpreted as corresponding to the available cardinal information. In this light, the set of all utility representations of the underlying preference system corresponds to the set of all cardinal scales compatible with the information encoded in the data. All analyses should be invariant under the specific representative. A prototypical example of such a situation are mutlidimensional spaces with differently scaled dimensions, where the information of the cardinal dimensions can only be used if it is ensured that this does not contradict the ordinal dimensions. We showed that many different tasks from applied  statistics can be information-efficiently framed in this way, ranging from fields as different as multi-dimensional poverty analysis, subgroup comparison in medicine, or credit risk comparison in finance. \\

But we did not want to content ourselves with the mere transfer of preference systems. Rather, we wanted to create a way of comparing random variables that take on values in preference systems in an information-efficient manner on the basis of the description we had obtained.\footnote{Since -- as already mentioned in the contribution --numerous challenges in statistics and machine learning can -- at least theoretically -- be broken down to comparing random variables, this seemed (and still seems) like a very reasonable goal.} The most natural idea for this was obvious: one could simply transfer the optimality criteria that we had developed for decision-making under weakly structured information to this statistical reinterpretation. Among these criteria, one seemed particularly natural: generalized stochastic dominance (GSD), as we have repeated, for example, in Definition~\ref{GSD} of the present thesis. It turned out that this did exactly what was needed. On the basis of GSD, we were able to define a meaningful, scale-robust and yet information-efficient \textit{stochastic ordering} between weakly structured random variables. Furthermore, we managed to make this ordering computationally accessible, to characterize it theoretically, to make it statistically testable as well as consistently estimable from data. Finally, we have also developed methods for the inference concepts based on GSD that can deal with deviations from the usual i.i.d. assumption. Given the generality of the problem addressed, namely the comparison of weakly structured random variables, further possible applications in various disciplines seem to be directly given. Examples range from (multidimensional) comparisons of risky assets by stochastic dominance (e.g.,   \citet{ll1984,whang2019econometric}), where GSD could also utilize the information encoded in potentially available cardinal dimensions, over the (multivariate) comparison of different treatment groups in medicine (see, e.g., \citet{leshno,davidov}), where GSD could be used for additionally accounting for the strength of the effects in the single dimensions, to multidimensional poverty analysis by means of stochastic dominance (e.g.,~\citet{lars,ARNDT20122290,foster}), where an GSD-based analysis could help to information-efficiently encounter for mixed-scaled dimensions such as income and formal education.\\

The last major step in this habilitation project was the transfer of the previous considerations to non-standard machine learning scenarios. The focus here was particularly on a general framework for conducting benchmark studies, which
\begin{itemize}
    \item can include several (potentially differently scaled) quality criteria \textit{simultaneously},
    \item adequately incorporate \textit{statistical uncertainty}, i.e. consider the benchmark suite used only as a sample of data sets, and
    \item can also assess potential inferential statements made about the quality comparison of algorithms with regard to their \textit{robustness} to the underlying assumptions of the respective inference concept.
\end{itemize}
Once again, the transfer of the GSD concept proved to be extremely fruitful in this context. The various quality metrics used can be used to set up a suitable preference system whose partial cardinal structure is generated by those metrics that have a cardinal scale level. The GSD relation based on this preference system then enables the multidimensional algorithm comparison, which satisfies all three desiderata mentioned above. This area also enables numerous exciting transfers to other areas of machine learning, as already discussed in detail in Section~\ref{frb} of the present thesis. \\

Finally, I would like to say the following: I hope that my work has made a (modest) contribution to understanding non-standard data and non-standard uncertainty situations in statistics and machine learning more as a possibility than an obstacle. If you look at these situations from the right perspective and have the right formalism at your disposal, you can master them without wasting information or having to make unjustified assumptions. The good news is that, for finding an adequate formalism, it may not be necessary to start from scratch: It is often enough to look beyond your own discipline to find what you are looking for. In my case, what I found was decision theory. I am convinced that the opportunities to benefit from this long-grown discipline are far from exhausted. \\

On the contrary, I have the feeling that the journey is only just beginning...
\newpage
\renewcommand{\bibname}{Further references}\markboth{Further references}{}
\addcontentsline{toc}{section}{References}
\bibliographystyle{abbrvnat}
\bibliography{literature}
\end{document}